\documentclass{article}

\usepackage[preprint]{neurips_2021}

\usepackage[utf8]{inputenc} 
\usepackage[T1]{fontenc}    
\usepackage{hyperref}       
\usepackage{url}            
\usepackage{booktabs}       
\usepackage{amsfonts}       
\usepackage{nicefrac}       
\usepackage{microtype}      
\usepackage{xcolor}         
\usepackage{amsmath}
\usepackage{bm}
\usepackage[normalem]{ulem}  

\usepackage[colorinlistoftodos]{todonotes}
\usepackage[labelfont=bf]{caption} 
\usepackage{subcaption} 
\usepackage[inline]{enumitem}

\usepackage[ruled,vlined]{algorithm2e}
\SetKwInOut{Break}{break}
\SetKwInOut{Parameter}{Parameter}
\usepackage[affil-it]{authblk}  
\usepackage[export]{adjustbox}
\usepackage{wrapfig}

\newcommand{\abs}[1]{\left\lvert #1 \right\rvert}

\title{Counterfactual Explanations via Latent Space Projection and Interpolation}

\author[1]{Brian Barr$^*$}  
\author[2]{Matthew R. Harrington$^*$}

\author[1]{Samuel Sharpe}
\author[1]{C. Bayan Bruss}

\affil[1]{\normalfont{Center for Machine Learning, Capital One}\protect\\
\textit{\{brian.barr, samuel.sharpe, bayan.bruss\}@capitalone.com}
\vspace{2mm}}

\affil[2]{\normalfont{Columbia University}\protect\\
\textit {m.harrington@columbia.edu}}

\begin{document}
\maketitle
\let\thefootnote\relax\footnotetext{$^*$Equal contribution.} 


\begin{abstract}
Counterfactual explanations represent the minimal change to a data sample that alters its predicted classification, typically from an unfavorable initial class to a desired target class. Counterfactuals help answer questions such as "what needs to change for this application to get accepted for a loan?". 
A number of recently proposed approaches to counterfactual generation give varying definitions of "plausible" counterfactuals and methods to generate them. However, many of these methods are computationally intensive and provide unconvincing explanations. 
Here we introduce \emph{SharpShooter}, a method for binary classification that starts by creating a projected version of the input that classifies as the target class.  Counterfactual candidates are then generated in latent space on the interpolation line between the input and its projection.
We then demonstrate that our framework translates core characteristics of a sample to its counterfactual through the use of learned representations. 
Furthermore, we show that SharpShooter is competitive across common quality metrics on tabular and image datasets while being three orders of magnitude faster than two comparable methods and excels at measures of realism, making it well-suited for high velocity machine learning applications which require timely explanations. 


\end{abstract}

\section{Introduction}
Machine learning models, leveraging ever larger datasets, have had notable recent successes in domains like healthcare, finance and industry.  These models exhibit well-known interpretability challenges. As a result, researchers have developed a variety of techniques to explain the model's behavior by providing feature importances for each data sample.  Some well known methods that provide post-hoc \textit{local explanations} include  those that observe deviations relative to a reference input (DeepLift~\cite{shrikumar2017} and Integrated Gradients)~\cite{sundararajan2017}, those that perturb the inputs and measure the effects (LIME)~\cite{ribeiro2016}, and those that take a game theoretic approach (SHAP)~\cite{lundberg2017}.  While each has its particular benefits and observed drawbacks they share a common goal: given a prediction, estimate the effect of each feature on that prediction. In a binary classification setting, these approaches seek to answer a similar question. For a model with two classes A and B and an input sample x, they attempt to answer the question "what evidence  did the model rely on to make its prediction of class A for input x?"  Answering this question, in this format, can be useful for an audience that is familiar with machine learning.

For other audiences, it may be useful to answer a slightly modified question "why didn't the model choose class B for this input?" and to get to that answer one might ask "what would have had to be different about x for it to result in B?" The exploration of outcomes in alternative but very similar worlds is known as counterfactual analysis. Here the alternative world is one in which the model had predicted something different. Given a trained model, the only way for the model to predict something different is for the inputs to change.  The counterfactual explanation is in the form of which features of the data have to change by what amount to alter the predicted class. There are many benefits to this form of explanation. Counterfactual reasoning is a common logical approach for humans and therefore is easy to interpret~\cite{fernandez2020}. 

However, counterfactual explanations are not without their limitations. Depending on the properties of the data, the classifier and the counterfactual method, it is possible to generate an out-of-sample counterfactual. Out-of-sample counterfactuals can result in the explanations not being feasible (i.e. unlikely to be attained since it is not seen in the training data). An example of this is adversarial samples~\cite{goodfellow2015explaining} which resemble the original class but have changed in imperceptible ways to fool the classifier. We note that for our purpose, valid counterfactual explanations are not adversarial in nature; they are a consequence of making meaningful changes to an input. In addition to challenges of remaining in-sample, counterfactuals are susceptible to a curse of dimensionality. High dimensional input spaces can produce high dimensional explanations with limited utility for those seeking to interpret the model (i.e they may not be sparse)~\cite{miller1956}. Additionally, searching over the high dimensional space of possible counterfactuals is computationally costly. 

The majority of the current methods focus on providing explanations to people impacted by the decisions made by these models.  Our work focuses on providing explanations to a neglected audience - the intermediaries ultimately responsible for their use, situated between model developers and the individuals that are subjected to the models decisions.  We focus on filling this gap with SharpShooter.  It is essential to provide those audiences with methods to guide  building, understanding, and  implementing models.  We emphasize their use as explanations for the model a necessary step before trying to afford recourse.

We require that a counterfactual explanation be high quality, feasible (i.e. in sample), sparse, and computationally efficient to generate.  To achieve these properties we propose SharpShooter a method for generating counterfactual explanations using latent representations. We demonstrate this approach on three datasets: MNIST~\cite{lecun98} (for its visual appeal) and two tabular datasets: the UCI credit dataset~\cite{uci_credit}  (important categoricals), and Lending Club~\cite{lendingclub} (less important categoricals). 

Our contributions are:
\begin{enumerate*}[label=(\roman*)]
    \item a model agnostic framework for finding actionable and realizable counterfactual explanations that is scalable in data with low computational cost
    \item a novel visualization of the decision boundary and the trajectories of the counterfactual search
    \item a comparison of methods across image and tabular datasets
\end{enumerate*}

\section{Background \& Previous Work}
This work sits at the intersection of generative modeling and explainable artificial intelligence (XAI).   We will detail some relevant work in the subsections below.
 
\paragraph{Generative modeling}
For SharpShooter, crafting counterfactuals requires creating new samples from a learned data distribution. Generative models, which attempt to learn $p(X)$ are well suited for the task.  Broadly speaking these approaches fall into two categories generative adversarial networks (GANs)~\cite{goodfellow2014generative} and normalizing flows~\cite{kobyzev2020normalizing,rezende2015variational}. Besides when used for improving the latent capacity of VAEs, both GANs and normalizing flows focus largely on generation (or $p(x))$, and less on the latent space of the hidden variables themselves.  For instance, GANs learn to generate synthetic samples from noise while normalizing flows use a series of transforms and inverse transforms to learn relationships between hidden variables and the input.

To gain greater control over the latent space we use variational autoencoders (VAE)~\cite{kingma2013auto}. Variational autoencoders have a similar encoder-decoder structure to autoencoders, however VAEs additionally learn a posterior distribution over latent codes through the use of variational inference~\cite{blei2017variational}. In practice VAEs are learned through maximizing a lower bound on $p(x)$ frequently known as the \textit{evidence lower bound} (ELBO)~\cite{doersch2021tutorial}. This formulation as a lower bound helps decompose the objective function into two parts: a reconstruction component related to the decoding network and a Kullback–Leibler divergence ($KL$) term that measures the asymmetric distance from the latent codes $p(Z|X)$ to the standard normal distribution of the prior $p(Z)$ in the standard VAE.  The normalization of latent space in VAEs through the $KL$ divergence helps provide structure over the latent codes $p(Z|X)$. In SharpShooter, we use the $\beta$-VAE to encourage continuous latent code spaces~\cite{higgins2016beta}. 

\paragraph{Explainable artificial intelligence} 

Much of the work on local explanations for machine learned models has focused on \emph{local attributions}, per sample feature importances and their aggregation~\cite{shrikumar2017,ribeiro2016,lundberg2017,datta2016,ibrahim2019}.  While useful for model developers, feature importances are less helpful in other contexts. The output and processes used to create these explanations are often difficult to understand for those not well versed with the background such as end-users.

A second branch of explainability is focused on counterfactual explanations; explanations that represent plausible changes for a user or observation to change their predicted classification. Counterfactual explanations are useful for multiple audiences, with most of the work centering on explanations for the people subject to a disadvantageous decision. Wachter et al~\cite{wachter2017} popularized the notion of using counterfactuals as way to provide recourse to \emph{data subjects} in response to the EU  General  Data  Protection  Regulation  (“GDPR”).  They leverage the principle of "the closest possible world" in motivating their work.  The notions of recourse~\cite{Ustun_2019} and fairness~\cite{joshi2019realistic} are still major themes of research.  This leaves a gap of providing counterfactual explanations for the audience that facilitates, mitigates and deliberates before a model gets deployed. 

A good survey of available methods is available in Verma et al~\cite{verma2020counterfactual}. Here we will frame much of that work through the lens of our use case, and highlight selected relevant work. We broadly separate the literature into two relevant branches, methods that involve input feature perturbations and those that use perturbations of hidden variables of generative or probabilistic models. We include comparisons to a method from each branch in our experiments in our experiments in Section~\ref{section:results}.

\paragraph{Input feature perturbations} One large branch of the literature creates counterfactuals by perturbing the input feature space.
For instance, early work~\cite{lash2016} under the name of \emph{inverse classification}  maintained sparsity by partitioning the features into immutable and mutable features, and imposing budgetary constraints on allowable changes to the mutable features. Likewise Laugel et al~\cite{laugel2018} advocated a sampling approach with their \textit{growing spheres} method to traverse the input space where Gomez et al~\cite{gomez2020vice} submitted a heuristic search method supported with visualisations to the FICO data challenge~\cite{explaina31}. Their considerations of data distribution and feature importance, along with counterfactual explanations are in line with our current work.

Dhurandhar et al~\cite{dhurandhar2018} utilize gradient descent in input space to find contrastive explanations,  which decompose into pertinent positives and pertinent negatives. They included an autoencoder loss to keep the explanations in sample. Van Looveren et al ~\cite{van_looveren2019} accelerate previous gradient descent methods with the introduction of \textit{prototypes}, guiding the gradient descent towards the average value of the target class as determined by averaging the training sets representation in the latent space. Russell~\cite{russell2019efficient} improves on the stability of the optimization problem posed in~\cite{wachter2017} and extends beyond considering only categorical features~\cite{Ustun_2019} in posing a mixed integer programming formulation. Where Poyiadzi et al~\cite{Poyiadzi_2020} create counterfactuals by forming a network of feasible paths following high density regions of input space.  GRACE~\cite{le2020grace} was designed for neural networks on tabular data that combines the concepts of contrastive explanations with interventions by performing constrained gradient descent adding an additional loss that is a measure of information gain to keep the resulting explanations sparse.

\paragraph{Latent space perturbations}
While intuitive in idea, input feature perturbation methods, without regularization, can generate counterfactuals unconvincing to the human eye that resemble traditional adversarial attacks~\cite{szegedy2013intriguing}.  To address this shortcoming, latent space perturbations methods utilize the machinery of generative and probabilistic models to insure counterfactuals have high probability under the data distribution $p(X)$.

For instance ExplainGAN~\cite{Samangouei_2018_ECCV}, a method for finding counterfactual explanations for images, jointly trains multiple autoencoders using the signal from the classifier and discriminators to inform the learned representations. Their approach is focused on deep learning and computer vision tasks, while our approach is domain and model agnostic. Pawelczyk et al~\cite{Pawelczyk_2020} focus on tabular datasets by perform growing sphere perturbation in the latent space of a conditional variational autoencoder to generate counterfactuals. Where Balasubramanian et al~\cite{balasubramanian2020latent} and Joshi et al~\cite{joshi2019realistic} directly perform gradient descent in the latent space of a variational autoencoder with various regularization terms. While other work focuses on topics of causality and counterfactuals within the context of generative models~\cite{mahajan2020preserving}.
 
Both latent space and input space methods differ substantially in the types of models and the datasets they can be used for, as well as the amount of model access they require (see Verma et al~\cite{verma2020counterfactual} for a full list). By contrast \emph{SharpShooter} is a model and data agnostic (although our primary focus is on tabular data).

\section{SharpShooter}
\begin{figure}[!hbtp]
\centering
    \includegraphics[height=2.0in]{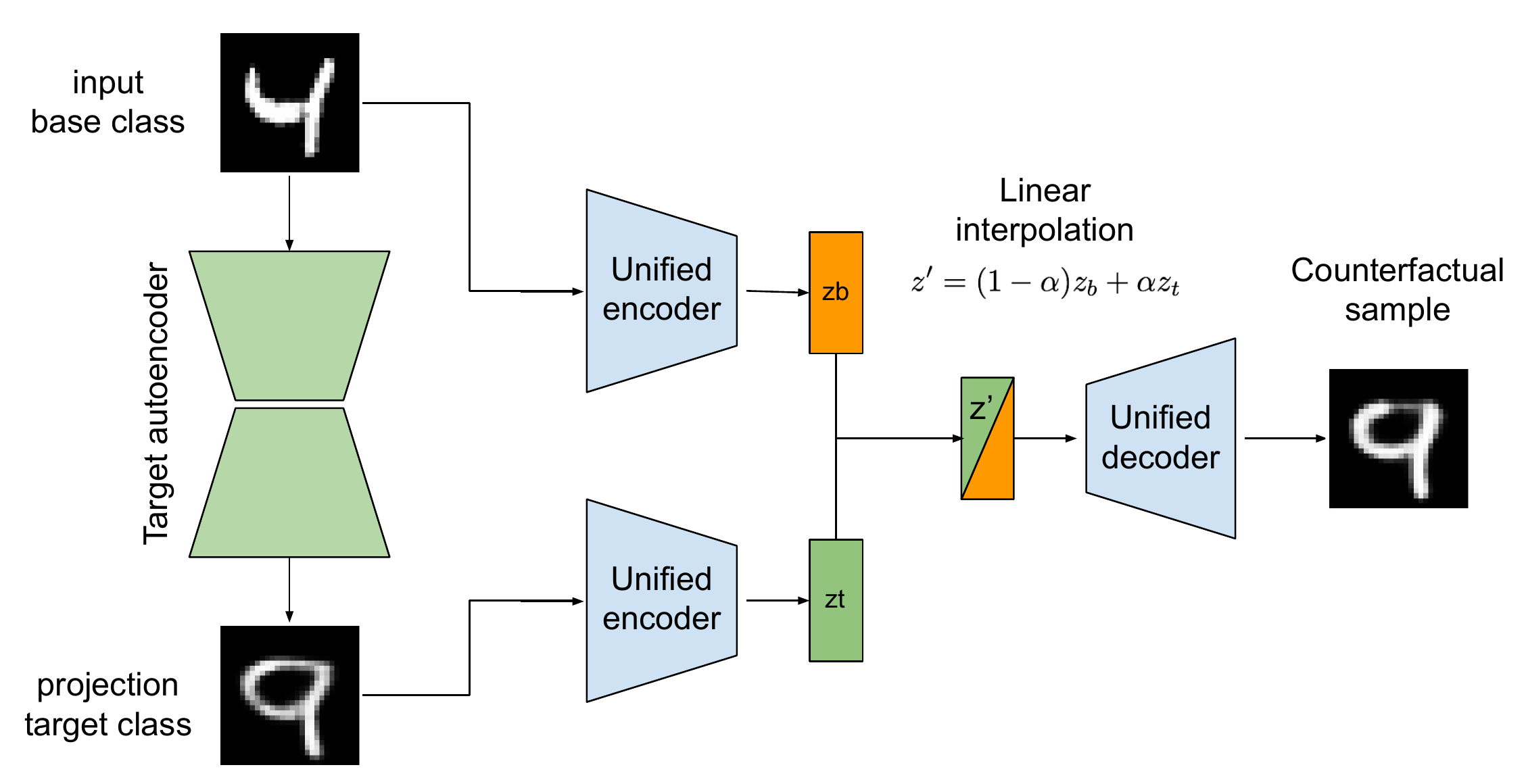}
    \caption{Diagram of the data flow in the SharpShooter algorithm. An input query sample, from the base class which receives a disadvantageous prediction from the classifier, is projected to target space via an autoencoder trained only on the target class.  Both the input sample and its projection are then embedded in a common latent representation learned by an autoencoder trained on both classes.  A linear combination of both latent codes, which when decoded crosses the decision boundary, results in a counterfactual sample.}
    \label{fig:ss_diagram}
\end{figure}

Our method relies on having access to the dataset and a previously trained classifier that the user wants to explain with counterfactual explanations. We will use the dataset to learn two distinct autoencoders with the algorithm flow shown in Figure~\ref{fig:ss_diagram}.

\paragraph{Target variational autoencoder} 

The starting point of the SharpShooter algorithm is the \emph{target} variational autoencoder (TVAE).  We train the TVAE on the subset of the data that is labeled as the preferred target class. A sample from the base class ($x_b$) is passed through the TVAE to give a reconstruction $\hat{x}_t$. This process creates a projection of the base observation to the target class.  

This raises an interesting question: how does an autoencoder make sense of data from a class it has never seen before?  If the model has only ever seen the target class, how does it reconstruct a base class sample? Does it place these latent codes in an area of very low data distribution and the sample becomes unrecognizable as the target class? or does the model generate valid reconstructions?

We find that with the use of the proper prior, the TVAE will correctly produce an in-sample projection. While too strong of a prior will collapse the latent distribution, a prior with wider variance prevents the creation of viable counterfactuals. This weighting is closely determined by the $KL$ coefficient $\beta$. A higher $\beta$ biases the model away from purely optimizing reconstruction loss and ensures that codes still have a roughly normally distributed shape.  This term ensures that even points with low mass in $p(x)$ have codes close to the prior distribution $p(z)$.  

The limited capacity of amortized inference models (such as a VAE) to both reconstruct and perform inference is known as the amortization gap~\cite{cremer2018inference}. In prior contexts, the amortization gap has been seen as an obstacle to overcome, but the SharpShooter algorithm takes advantage of the lossy gap in inference when translating bases to target counterfactuals. If reconstruction and latent code inference becomes too direct then predictions on the unseen class stop classifying as the target class. We show this trade-off and investigate it in the appendix and in Figures~\ref{fig:mnist_grid_search_tvae}-\ref{fig:lc_uvae_grid_search}.

\paragraph{Unified variational autoencoder} 
We train the \emph{unified} variational autoencoder (UVAE) on both classes.  The encoder of the UVAE is used to embed both a base class sample, $z_b$,  and its target class projection, $z_t$, in a common learned representation. We then use linear interpolation to sample along the line between a base sample and its TVAE-generated projection in UVAE space. We use the UVAE to estimate the joint data distribution and its latent space so that we can ensure counterfactuals both lie on the data manifold, and that they are close to their original sample in latent distance. While a powerful method that seems to succeed in our experiments on MNIST and the Lending Club datasets, there are additional difficulties building non-degenerate $\beta$-VAEs on the highly imbalanced UCI Credit dataset with its higher tabular dimensionality.


\subsection{Algorithm}
Algorithm~\ref{algo:SA} shows pseudo code as a companion to Figure~\ref{fig:ss_diagram} which is used to generate the counterfactuals in Section~\ref{sec:results}.

\begin{algorithm}[h]
\Parameter{$p$ probability of target counterfactual class (0.5 for decision boundary), $tol$ tolerance}
\KwIn{Classifier $f$, VAE trained on $X_t$ (TVAE), VAE trained on $(X_b, X_t)$ (UVAE), $S$ samples $\bm{\alpha} =\alpha_{s=1}^S \> \in (0,1)$} 
\KwOut{$X_{cf}$ a counterfactual}

$\hat{X}_t  = \phi_t(X_b)$ \\  
$Z_b = \varphi_{b,t}(X_b)$ \\  
$\hat{Z}_t = \varphi_{b,t}(\hat{X}_t)$ \\

\For{$\alpha_s$ in $\bm{\alpha}$}{
    $Z'(\alpha_s) = \alpha_s \hat{Z}_t + (1-\alpha_s) Z_b$ \\
    $X' = \psi_{b,t}(Z')$

    \If{$\abs{f(X') - T < tol}$ \textbf{and} $f(X') > p$}{
        $X_{cf} = X'$ \\
        \Return $X_{cf}$}
    }

\caption{SharpShooter; $\> \alpha$-\emph{Sampled} }
\label{algo:SA}
\end{algorithm}

The dataset is composed $\mathcal{D} = (y_i, X_i)^n_{i=1}$ where we can split $X$ into two classes $X = (X_t, X_b)$ where $X_{t,i}$ has $\hat{y}_i = 1$ and $X_{b,i}$ has $\hat{y}_b = 0$ under the classifier $\hat{y}= f(X)$. For training we separate the dataset according to the class and train the target VAE $\phi_t$ and the unified VAE $\psi_{b,t}$.  We denote their encoders $\varphi_t$ and $\varphi_{b,t}$, and their decoders $\psi_t$ and $\psi_{b,t}$ respectively.

We first pass a base sample $x_b$ through the TVAE $\phi_t$ to obtain the sample's target projection $\hat{x}_t$. Next, we take both $x_b$ and $\hat{x}_t$ and pass them through the encoder of the UVAE $\varphi_{b,t}$ to obtain $z_b$ and $\hat{z}_t$ respectively. Once we have both "endpoints" in joint latent space, an encoded counterfactual candidate, $z'$, is generated using linear interpolation \begin{math} z' = (1 - \alpha) z_b + \alpha z_t\end{math} with $\alpha$ in $(0,1]$. The candidate counterfactual is decoded and the classifier assesses its fitness.  The counterfactual is then accepted if it both crosses the decision boundary and is within a user specified tolerance $tol$ which is the desired maximum distance from $T$ for the generated counterfactual's classification score. For the experiments in Section~\ref{section:results}, the decision boundary $T$ is set to be $p(x) = 0.5$.

The line search is then executed by sampling a finite number of $\alpha$.  An alternative formulation of this search can also be done using a one-dimensional gradient descent which is demonstrated in Algorithm \ref{algo:GD} in the appendix. Gradient descent can be advantageous over a simple line search depending on the user criteria (tighter tolerance can be hard to hit with sampling) and the use case (the presence of categoricals can make the probabilities along the line either discontinuous and/or non-monotonic).

\subsection{Visualization of the latent space, and decision boundary}

\begin{figure*}[ht!]
    \centering
    \begin{subfigure}{0.4\textwidth}
        \centering
        \includegraphics[width=.98\linewidth]{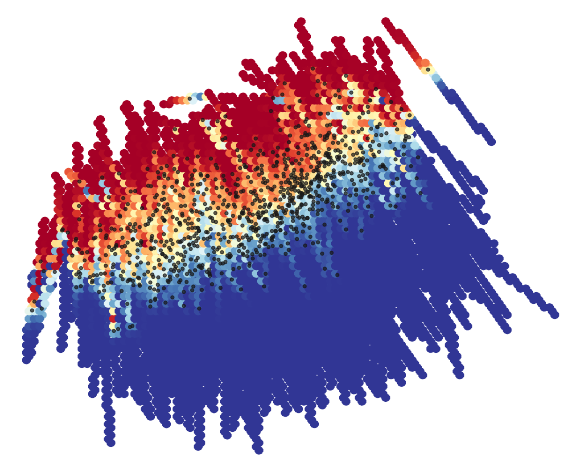}
        \caption{MNIST}
        \label{fig:hex_a}
    \end{subfigure}
    \begin{subfigure}{0.4\textwidth}
        \centering
        \includegraphics[width=.98\linewidth]{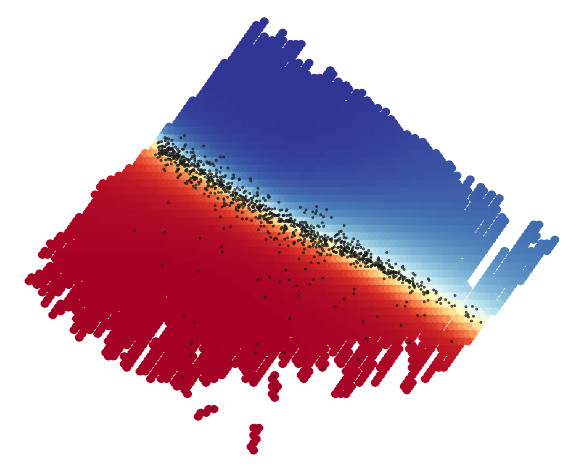}
        \caption{Lending Club}
        \label{fig:hex_b}
    \end{subfigure}
    \caption{Visualization of the decision boundary in UVAE latent space, with classifier prediction values being drawn from samples along the interpolation line between the encoding of the base class and their SharpShooter candidates. The black points represent the location of the chosen counterfactual samples. Blue areas represent regions of high base classification where red represents regions of high target class classification.} 
    \label{fig:decision_boundary}
\end{figure*}

We visualize the decision boundary in Figure~\ref{fig:decision_boundary} using the low dimensional projections of the latent codes in Unified VAE space for the base class $Z_b$, the target class $Z_t$, and the generated counterfactuals $Z_{cf}$. The colors denote prediction score (with blue denoting the base class) and the black dots represent the final SharpShooter counterfactuals.

The visualization of the decision boundary in Figure~\ref{fig:decision_boundary} was generated by sampling many points along the candidate counterfactual line $Z'(\alpha_s) = \alpha_s \hat{Z}_t + (1-\alpha_s) Z_b$. The interpolated points are then decoded by the UVAE and passed to the classifier to obtain a classification score: i.e. $f(\psi_{b,t}(Z'))$.  Plotting is then done using a weighted hexbin over the UVAE space to plot in two dimensions. Figure~\ref{fig:tsnes} shows the underlying codes for all classes and their projections, and Figure~\ref{fig:decision_bound_lines} visualizes the candidate counterfactual lines pairing $z_b$ with their corresponding $\hat{z}_t$ with more accompanying description found in Appendix \ref{app:vis}.

Figure~\ref{fig:decision_boundary} offers valuable insight into both the topological structure of the classifier's decision boundary and the Unified VAE's latent space. In the case of the Lending Club dataset, we see two clearly defined regions of classification scores. Without labels the UVAE is managing to create an approximately linear separation between the two classes with a highly nonlinear transformation mapping from code to covariate space. Even in Figure~\ref{fig:hex_a}, where we only plot the first two PCA components of a 20-dimensional latent space, we see a boundary between the two classes with large regions of classifier probability close to 0 or 1. While the boundary for MNIST is fuzzier than in Figure~\ref{fig:hex_b}, this effect is largely due to greater dimensionality of MNIST's latent code space than in the UVAE for Lending Club. With a few notable exceptions, the two classes are separable by a single hyperplane. These results indicate an exciting correspondence between the classifier and the UVAE, even though the two models have disparate objective functions and differ in their architectures.

Sampling along the interpolation line from a base to its transformed counterfactual provides an attractive path to visualize the qualities of the classifier to provide simplified views of an abstract space. By comparison, performing a similar exercise, but instead sampling an evenly spaced cloud of points in $\varphi_{b,t}$ space will place too much weight on low-probability points. Likewise, as we are travelling along the same trajectory that the algorithm itself traverses when selecting the chosen counterfactual this visualization method can provide further intuition into the workings of the SharpShooter process.



\section{Results}
\label{sec:results}
We compare SharpShooter (SS) against two other standard counterfactual methods: gradient descent in input space (GDI) and gradient descent in latent space (GDL). In both cases gradients are taken of the loss function of the distance from the classifier score to the chosen cutoff $T$. GDI is then solved by making updates to the input until the distance is less than the tolerance $tol$, that is, $\min_{X_b} (T-f(X_b))^2$, whereas GDL is given by making updates to the latent representation $Z_b = \varphi_{b,t}(X_b)$, that is, $\min_Z (T-f( \psi_{b,t}(Z) ))^2$. Note that while a few other counterfactual methods exist that are model agnostic and work on both tabular and image datasets, GDI and GDL broadly categorize many of the alternative approaches taken in the literature. 

\paragraph{Metrics for comparison} 
\label{metrics_section}
We examine the quality of counterfactuals by tabulating these metrics surveyed from literature:  
\begin{enumerate*}[label=(\roman*)]
    \item time required to find a CF for a given sample
  \item validity (val), the percentage success in generating counterfactuals that cross the decision boundary
  \item proximity (prox), distance from input sample to counterfactual in latent space
  \item sparsity (spars), percentage of features changed
  \item classifier shift - pass CF sample through UAE and check the change in classification score
  \item reconstruction loss - a measure of the CF being in sample.
\end{enumerate*}
Further details about metrics and the classifiers used in the experiments can be found in the appendix.

\label{section:results}
\begin{figure}[hbt!]
    \centering
    \adjincludegraphics[width=.85\columnwidth,trim=0 {.1\height} 0 {.125\height},clip]{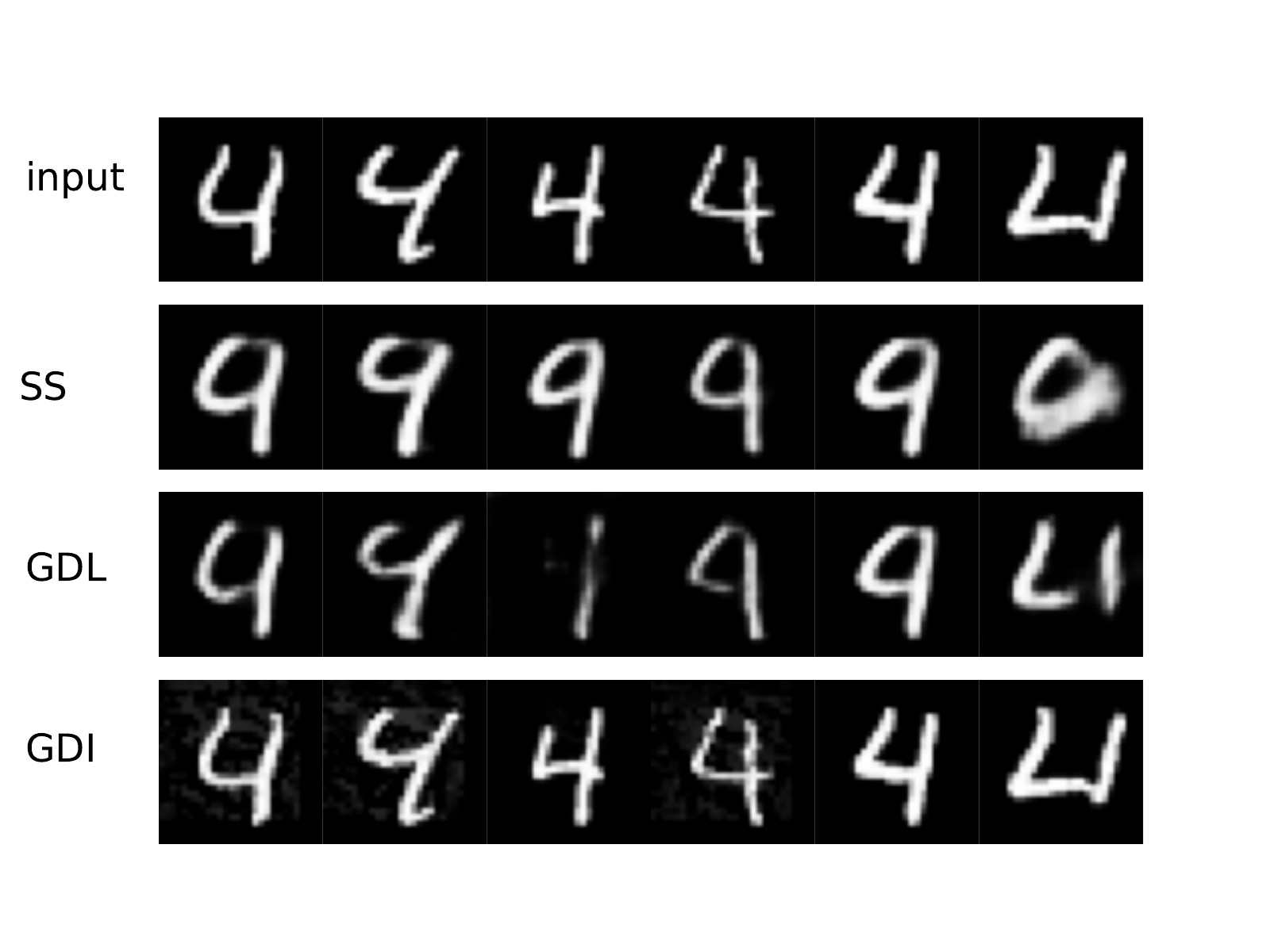}
    \caption{Input and counterfactual samples for MNIST showing from top to bottom: original input image ($4$), and counterfactuals found via SS, GDL, and GDI. Note the SharpShooter examples exhibit more of the characteristics of the target class ($9$) than the other methods.}
    \label{fig:mnist_samples}
\end{figure}
\paragraph{MNIST}

For MNIST, we formulate the problem as a binary classification problem of predicting fours (our base class) and nines (our target class). While multiclass image counterfactuals are not the primary focus of SharpShooter, MNIST provides a naturally intuitive visualization of the impact of projecting from the base class to the target class.  The counterfactual examples from SharpShooter shown in Figure~\ref{fig:mnist_samples} are exhibit  a blend of features from the input (e.g. angle of strokes) and from the target class (e.g. rounder and more closed tops). Interestingly, all methods show struggles in interpreting the last out-of-sample image of a four.

The quality of counterfactual explanations as measured by the metrics discussed in the appendix are summarized in Table~\ref{mnist-metrics-table}. These metrics show SharpShooter outperforming the other methods in validity, classifier shift, reconstruction score, and time, but lagging in proximity and sparsity.

Note that the thin layer of changed pixels in the counterfactuals for GDI in Figure~\ref{fig:mnist_samples} is reflected by the method's poor performance in classifier shift and reconstruction score. The noisy transformations seen in those counterfactuals is not reproduced in UVAE reconstructions - leading to the lower performance in CS and RECON. This is why we introduce these two measures as metrics of realism of counterfactuals: the UVAE smooths out samples far from the underlying data distribution. Meanwhile, SharpShooter counterfactuals lie closer to the data distribution, and appear more like the target class in Figure~\ref{fig:mnist_samples}.


\begin{table}[!htpb]
  \caption{Summary of metrics for MNIST digit recognition}
  \label{mnist-metrics-table}
  \centering
  \begin{tabular}{l|l|l|l|l|l|l|l|l}
    Method  &   val  &  prox  &   spars  &   CS    & recon  & time\\ 
    \midrule
    SS      &  \textbf{0.97}  &  1.36  &   0.32             & \textbf{0.13}  &  \textbf{1.5e-3} & \textbf{3e-3}  \\ 
    GDL     &  0.79           &  1.13  &   \textbf{0.30}    & 0.17           &  3.3e-3          & 1.82  \\ 
    GDI     &  0.67  &  \textbf{0.84}  &    0.42            & 0.41           &  6.4e-3          & 1.81  \\ 
      
    \bottomrule
  \end{tabular}
\end{table}

\paragraph{UCI Credit Card Default}
Following new banking regulation in 1990~\cite{wang}, Taiwan faced a credit card debt crisis. The UCI credit default dataset~\cite{uci_credit}, contains demographic, billing and payment information and labels of default from the period of April to September of 2005 while the crisis is deepening.  The UCI credit dataset contains a high number of categorical variables. Complicating things further, the labels have a large class imbalance on the scale of 20 to 1 in favor of the non-defaulted class. A classifier is built to predict default with an AUC of 0.76 - one of the drawbacks of modeling an unsteady temporal process as steady, there is noise on the labels in the form of samples that have yet to default.


Due to these complications, training a standard VAE for either part of the SharpShooter process proves more difficult than training for the other datasets.  The nature of $KL$ divergence attempts to form spherical latent distributions and is zero-avoiding\cite{dieng2019avoiding}, while at the same time the reconstruction loss on the categorical variables encourages latent codes to be further apart. The tension caused by these competing forces can cause for either collapsed latent spaces or areas of low probability mass. This push and pull can be seen in the TSNE of UCI credit's latent codes (see Figure~\ref{fig:tsnes}) which show islands of mass, even within the same class type hinting at clusters in the latent space. The shape of the latent space for the TVAE and UVAE thus requires careful tuning parameters such as $\beta$ and categorical reconstruction weights. 

The UCI credit dataset serves as a good counterexample necessitating the use of more expressive generational models with fundamentally different code distributions $p(z|x)$ for tabular datasets with higher categorical dimensionality.

\begin{table}[!htpb]
  \caption{Summary of metrics for UCI credit default}
  \label{uci-metrics-table}
  \centering
    \begin{tabular}{l|l|l|l|l|l|l|l|l}
        Method  &   val            &  prox          & spars         &   CS         & recon    &    time\\ 
        \midrule
        SS      &  0.505           &  3.99          &  0.309       & \textbf{0.136} &  1.58 & \textbf{1e-3}  \\ 
        GDL     &  0.317           &  2.25          &    0.113       & 0.154        &  0.74 &  8.14  \\ 
        GDI     &  \textbf{0.752}  & \textbf{0.83}  &\textbf{0.000}  & 0.170        &  1.28  & 8.02  \\ 
        \bottomrule
      \end{tabular}
\end{table}

\begin{figure}[!htpb]
    \centering
    \includegraphics[width=.95\linewidth]{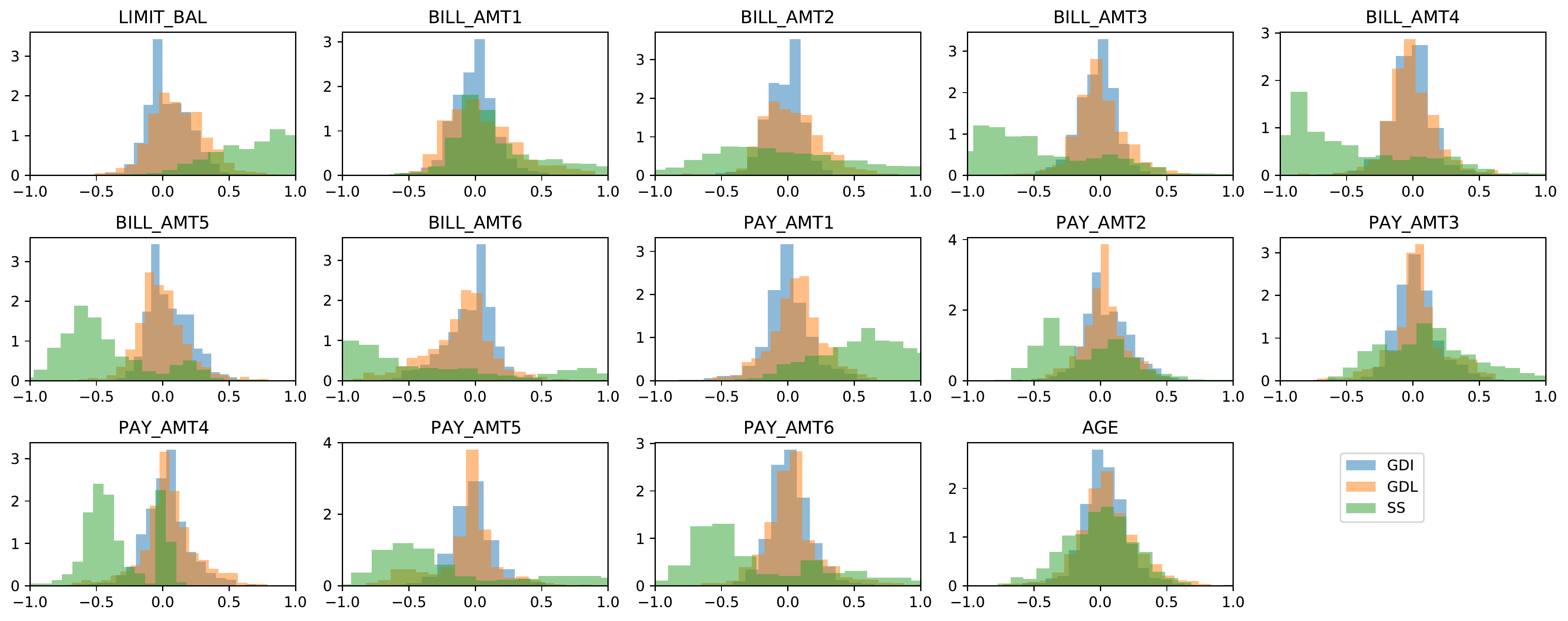}
    \caption{Distribution of counterfactual changes made for the continuous variables in the UCI dataset. The changes SharpShooter (SS) make highlight the impact of lower spending and matched lower payment.}
    \label{fig:uci_cont}
\end{figure}

Changes made by the three methods on the continuous variables of the dataset - presented in a scaled log value - can be seen in Figure~\ref{fig:uci_cont}.  The SharpShooter changes in bill amount are in striking contrast to the gradient descent methods.  While similar for April, a structural change is seen in May, and the later months suggest drastic reductions in bill amounts.  The reductions suggested in payment amount correlate with lower bill amounts.

\begin{figure}[!htpb]
    \centering
    \adjincludegraphics[width=.8\linewidth, trim=0 0 0 {0.08\height}]{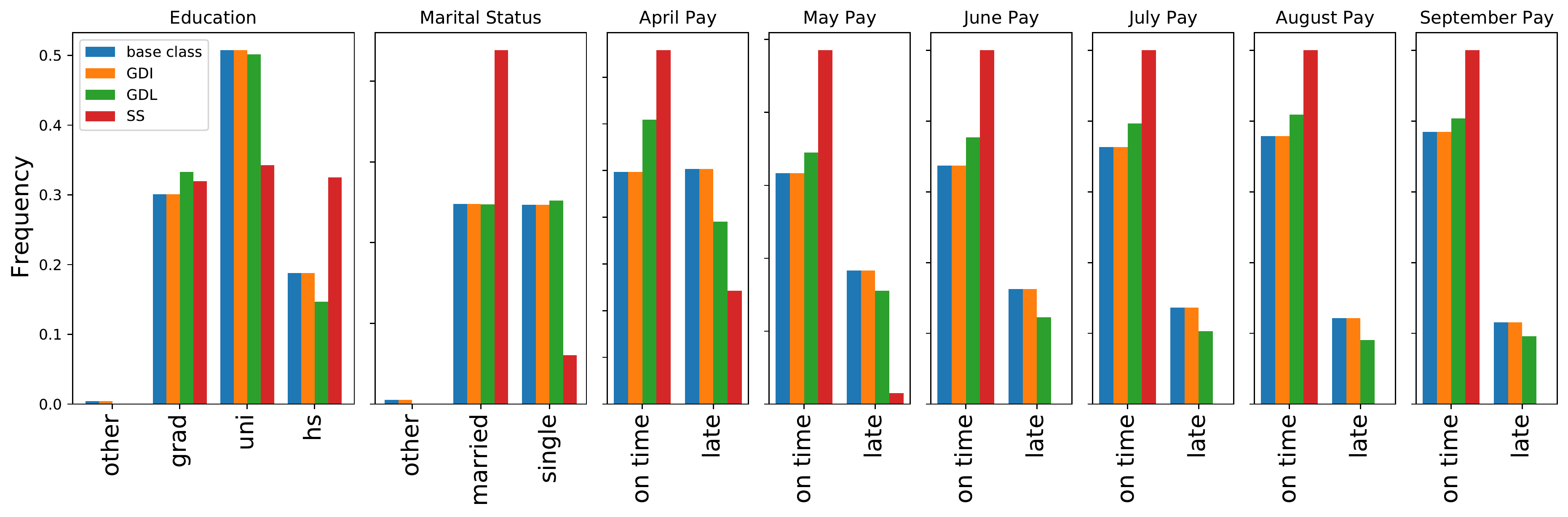}
    \caption{Summary of categorical features for counterfactual explanations compared to the base class. While the two gradient descent methods rarely change categorical values, SharpShooter has a higher likelihood of finding counterfactuals that pay on time and get married.}
    \label{fig:uci_categoricals}
\end{figure}

Likewise Figure~\ref{fig:uci_categoricals} shows values for education, marital status, and pay timing. For categorical variables, SharpShooter shows a tendency to create counterfactuals where the payment is on time, and the marital status changes to married. On the other hand, GDI and GDL tend not to make many changes to categorical variables.  This could be due to a relative inability of gradient descent methods to change categorical variables, a property not shared by the VAE-based SharpShooter.

\paragraph{Lending Club}
Lending Club is a peer-to-peer lending company that would periodically release data on its loan portfolio. This tabular dataset included information on whether a borrower defaulted on their loan, the size of the loan, their annual income, debt to income ratio, FICO score, and the loan length. We train a classifier to predict default using five continuous and one categorical feature and obtains an AUC of 0.95.  A summary of quality metrics for the methods is seen in Table~\ref{tab:lc_metrics}.

\begin{table}[!htpb]
  \caption{Summary of metrics for Lending Club loan default}
  \label{lc-metrics-table}
  \centering
  \begin{tabular}{l|l|l|l|l|l|l|l|l}
    Method  &   val   &  prox  & spars  &   CS    &  recon  & time\\ 
    \midrule
    SS      &  \textbf{0.98} &  1.66          &  \textbf{0.49}  & 0.33          &  0.019    & \textbf{1.6e-3}  \\ 
    GDL     &  0.91         &  1.35           &  0.74           &  0.313        &  \textbf{0.015} & 0.71  \\ 
    GDI     &  0.93         &  \textbf{0.71}  &  0.50           & \textbf{0.19} &  0.037          & 1.21  \\ 
    \bottomrule
  \end{tabular}
  \label{tab:lc_metrics}
\end{table}

For this experiment, the latent dimension of both autoencoders is two.  In low dimensional space, the reconstruction loss performance of SharpShooter becomes very close to GDL. Otherwise Sharpshooter shows high validity and generates counterfactuals that are more sparse. Even though SharpShooter sees a larger classifier shift and reconstruction loss on this dataset, time is still much lower than the other two methods.

\begin{figure}[!htpb]
    \centering
    \adjincludegraphics[width=.95\linewidth, trim=0 0 0 {0.05\height}, ]{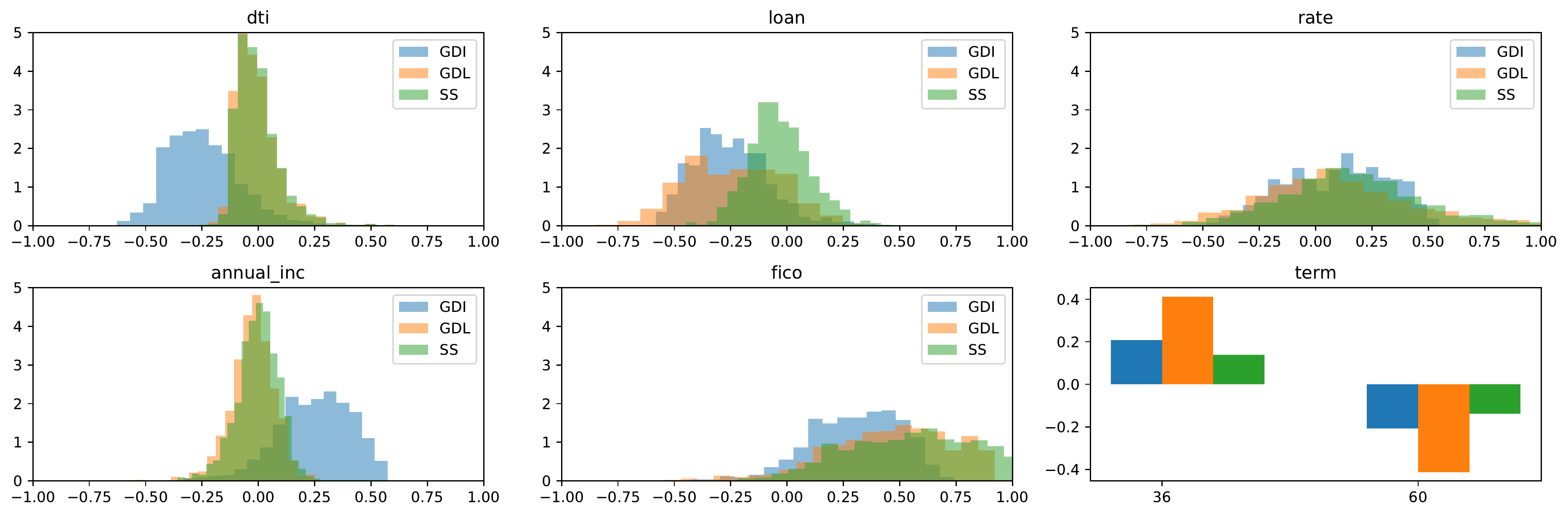}
    \caption{Distribution of changes made for Lending Club counterfactuals.  Gradient descent in input space makes the most extreme changes, while SharpShooter in the two dimensional latent space performs finds similar explanations as gradient descent in latent space.}
    \label{fig:lc_cont}
\end{figure}

To illustrate these differences between methods the distributions of changes made for counterfactual explanations are shown in Figure~\ref{fig:lc_cont}. Given the latent search space was only 2, it is unsurprising that SharpShooter and GDL were not very large. The most notable difference between the two sets of counterfactuals was with respect to loan size, which was mostly unchanged in SharpShooter counterfactuals. Meanwhile, gradient descent in the input space makes the largest changes for variable like debt to income ratio (dti) and annual income. 

\section{Conclusion}
In this paper we present SharpShooter: a model-agnostic algorithm for finding counterfactuals via linear interpolation in latent space. SharpShooter implements a framework that uses two variational autoencoders and introduces the concept of a target VAE that projects a sample to a counterfactual class. We demonstrate SharpShooter's advantages and disadvantages when compared against two general methods on three datasets (MNIST,  UCI credit card default, and Lending Club loan default). We show that while not always the most sparse or proximal, our method is orders of magnitude faster on all datasets and excels at realism, as measured by two novel metrics. We provide a thorough examination of the created counterfactuals and their attributes, meanwhile suggesting that future work could revolve around improving latent space structure. Lastly we introduce a straightforward method of visualization of the decision boundary that helps provide insight into the classifier, the joint representation space, and the SharpShooter process.



\bibliography{cf}

\appendix
\section{Appendix}

\subsection{Description of Comparison Metrics}
Comparison metrics used are a combination of commonly used metrics from the literature and two novel comparison methods, explained below. Except in the case of validity, all metrics are only calculated for the subset of counterfactual samples that classify in the correct target class. 

\paragraph{Validity}
Validity is the percentage of generated counterfactuals that correctly cross the decision boundary.  The main goal for generating counterfactuals is that they are, indeed, counterfactuals, which requires them to classify as the target class. We take validity with respect to the classifier, noting this makes our practice distinct from analyzing counterfactuals in the causality literature. 

\paragraph{Proximity}
Proximity is discussed in many papers that promote algorithms for counterfactual generation \cite{verma2020counterfactual}. This metric measures how close a counterfactual is to to the base sample it is generated from.  Typically this measure is taken as the $L_2$ norm in input space.  Here we use the L2 norm in representation space, although we note similar (unreported) patterns for L2 in input space.  We do note that proximity is limited in that it does not necessarily tell us much about whether the generated counterfactuals are actually on the data manifold (as seen in Figure~\ref{fig:mnist_samples}).  Hence we also report two novel measures \emph{classifier shift} and \emph{reconstruction loss}, explained below. 

\paragraph{Sparsity}
Sparsity is another commonly used metric in the counterfactual literature \cite{verma2020counterfactual}. Sparsity is a measure of how sparse the change vector is. In our usage, we calculate the change vector and then we count the percentage of dimensions that see changes greater than a half pixel in MNIST, and that see non-zero changes in the categorical and continuous variables in the two tabular datasets.

\paragraph{Classifier Shift}
Classifier shift and reconstruction loss are two novel measures of whether a generated counterfactual is \emph{realistic}. These measures bear resemblance to the VAE-specific measures of $IM1$ and $IM2$ promoted by Van Looveren and Klaise \cite{van_looveren2019}. Classifier shift is given by $|f(x_{cf}) - f(\phi_{b,t}(x_cf)|$ where $f$ is the classifier and $\phi_{b,t}$ is the UVAE. In other words, we check how much passing the given counterfactual through the UVAE changes its original classification score.

The classifier shift metric relies on the strength of the prior to pull samples that lie farther from the data manifold closer to a higher probability reconstruction.  If a sample does not have high probability, the resulting reconstruction from the UVAE will see a large amount of change in terms of its classification score, which is reported in the tables above. 

\paragraph{Reconstruction Loss}
Similarly to classifier shift, another way in which we can measure the stability of a given counterfactual is to measure its reconstruction loss.  The reconstruction loss measures the change in the evidence lower bound (ELBO) for the log likelihood of a sample $x$. Hence we can measure the effect of passing a counterfactual through the UVAE also by measuring its loss with respect to the UVAEs reconstruction (i.e. $-ELBO_{\phi_{b,t}}(x_{cf})$). Having a smaller loss is desirable here.

\paragraph{Time}
Time, one of the largest benefits of the SharpShooter process, here is reported as the time it takes to generate a single counterfactual sample, on average. SharpShooter is particularly fast in this dimension compared to other methods because generating a sample only requires the predictions of two variational autoencoders for a limited number of points for a fixed number of interpolated points. By comparison, iterative solutions based on gradients or perturbing the input space can be quite expensive if the search distance is large. Notably, we here exclude the overhead time of training the VAEs because, while this time is not trivial, it is exclusively a fixed cost and is not related to the scalability of generation of counterfactuals for a sample of size $n$.

\subsection{Visualizations of SharpShooter Latent Spaces}
\label{app:vis}

\begin{figure*}[ht!]
    \centering
    \begin{subfigure}{0.3\textwidth}
        \centering
        \adjincludegraphics[width=.98\linewidth]{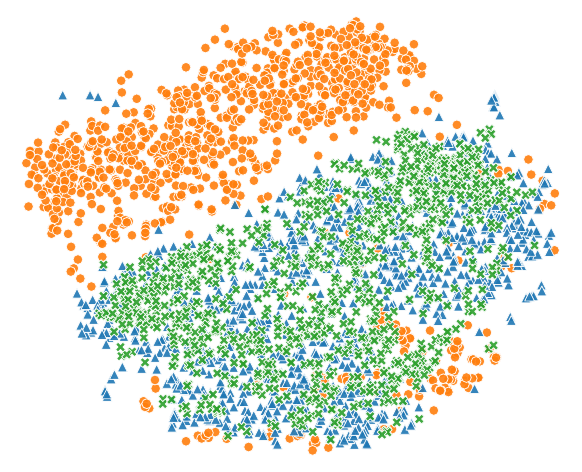}
        \caption{MNIST}
    \end{subfigure}
    \begin{subfigure}{0.3\textwidth}
        \centering
        \adjincludegraphics[width=.98\linewidth]{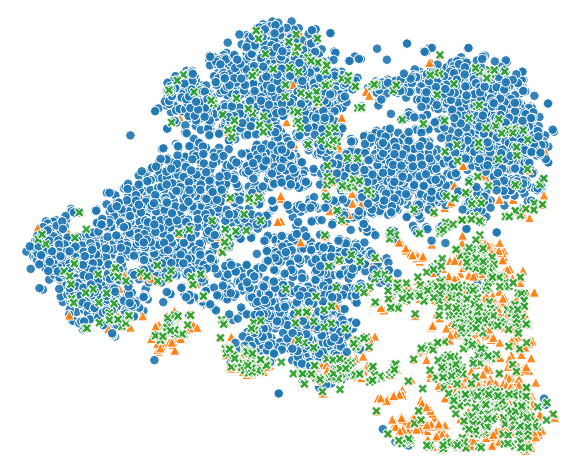}
        \caption{UCI credit}
    \end{subfigure}
    \begin{subfigure}{0.3\textwidth}
        \centering
        \adjincludegraphics[width=.98\linewidth]{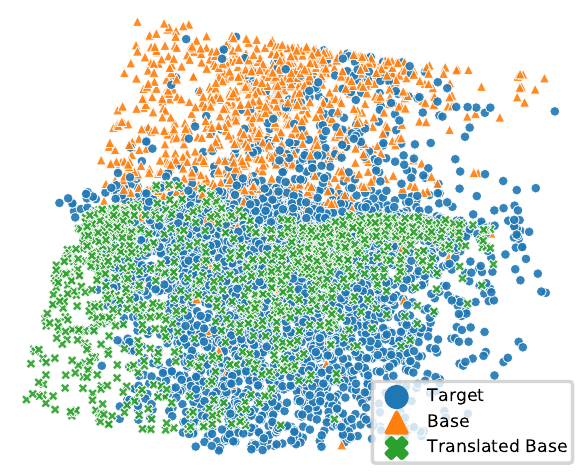}
        \caption{Lending Club}
    \end{subfigure}
    \caption{(a) and (c) show the first two dimensions of a PCA plot of the latent codes in Unified VAE space of the test datasets for their respective datasets. (b) shows the codes after a t-distributed stochastic neighbor embedding (TSNE) transformation. We plot the codes of base class, target class, and projections.} 
    \label{fig:tsnes}
\end{figure*}

\begin{figure*}[ht!]
    \centering
    \begin{subfigure}{0.49\textwidth}
        \centering
        \includegraphics{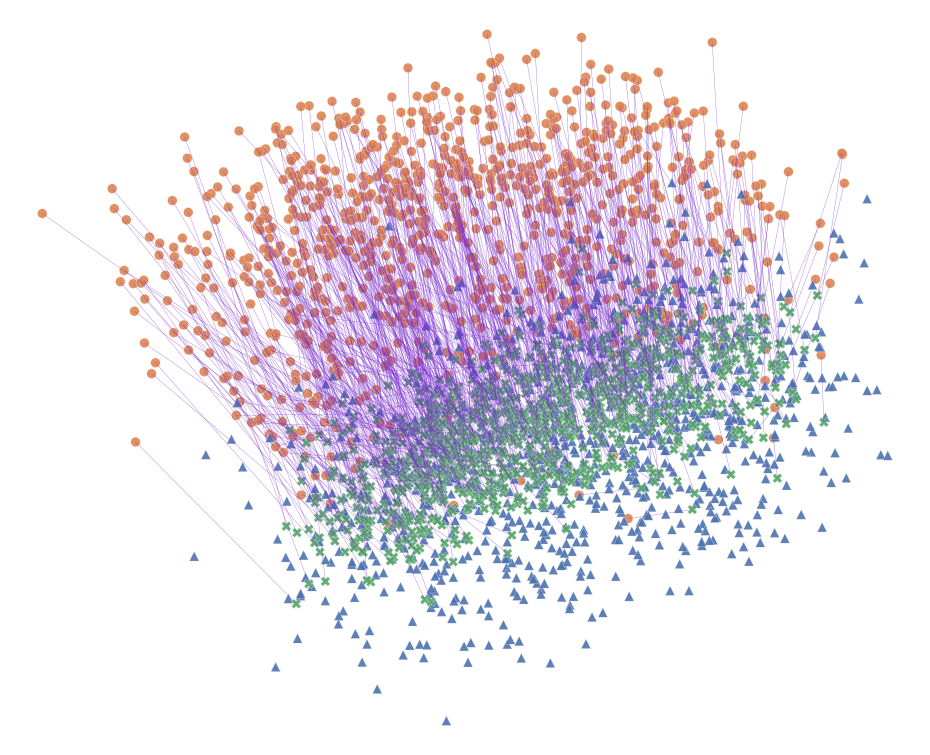}
        \caption{MNIST}
    \end{subfigure}
    \begin{subfigure}{0.49\textwidth}
        \centering
        \includegraphics{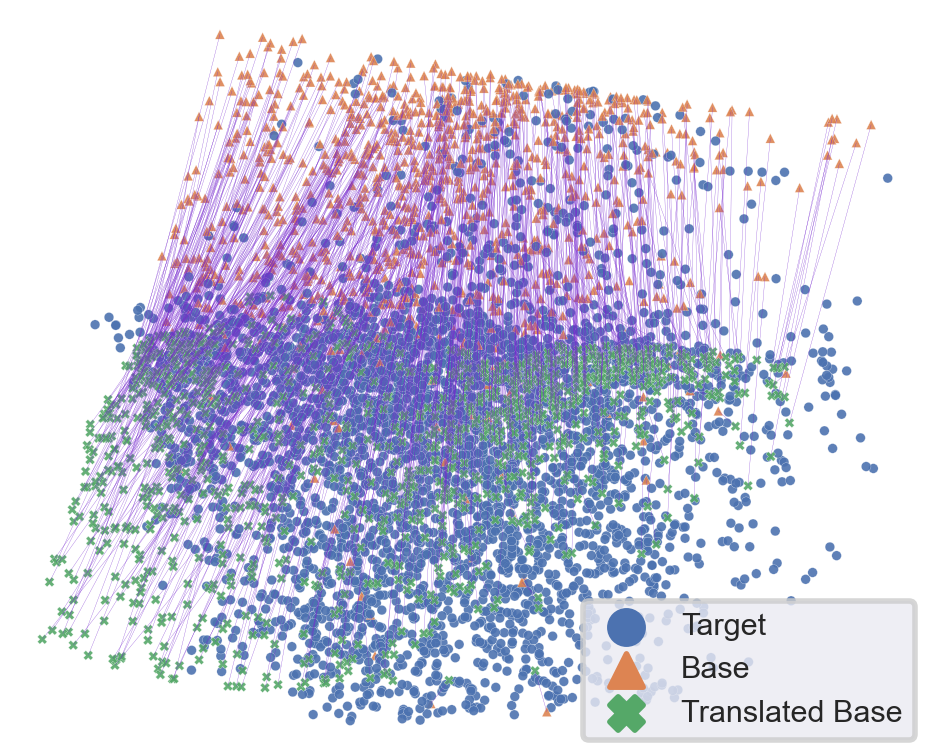}
        \caption{Lending Club}
    \end{subfigure}
    
    \caption{Showing the same projection of codes in Figure \ref{fig:tsnes} (a) and (c), we plot lines connecting the original base observations and their Target VAE projections in the latent space of the Unified VAE. Candidate counterfactuals are then sampled as linearly interpolated codes from points along these lines.} 
    \label{fig:decision_bound_lines}
\end{figure*}

Figure \ref{fig:tsnes} and Figure \ref{fig:decision_bound_lines} give insights into how SharpShooter generates counterfactuals and how visualizations were created in Figure \ref{fig:decision_boundary}.

In Figure \ref{fig:tsnes} we first plot the projections of all classes (base $Z_b$, target $Z_t$, and translated base $\hat{Z}_t$) in Unified VAE (UVAE) space.  This plot shows distinct separation of $Z_b$ and $Z_t$ in the cases of MNIST and Lending Club with some minor exceptions. Likewise for these datasets, $\hat{Z}_t$ seem well mixed in with the target class as desired.  This is the intended outcome of the TVAE's projection for SharpShooter to produce meaningful counterfactuals because we need a base's projection to be classified as a target for linear interpolation to cross the decision boundary. By comparison, the higher dimensional, more heavily imbalanced UCI dataset shows less clear separation of $Z_b$ and $Z_t$, with $\hat{Z}_t$ jointly mixed amongst both base and target codes.

Figure \ref{fig:decision_bound_lines} now examines exactly which point in $Z_b$ is paired with which point in $\hat{Z}_t$. I.e. now we plot a line from a base point to its TVAE transformation.  The most immediate observation from this figure is the similar angle and length of many of the lines.  This observation is not unfamiliar to the representation learning literature of similar difference vectors between two points implying similar changes. For instance, the classic example is taking the difference vector between "Italy" and "Rome" and adding it to "France" to produce "Paris". This observation raises an interesting question of to what extent can counterfactual be interpreted as a common vector in some representation space.

Lastly, Figure \ref{fig:decision_boundary} is produced by taking a large number of $\alpha$ to produce many $Z'$, decoding those interpolated codes, and calculating each point's corresponding classification score. Finally, we plot this large cloud of points as a weighted 2d histogram, and in this case we use a weighted hexbin plot to avoid visual biases. We here note that while the Lending Club dataset is an actual plot, the MNIST plot is much higher dimension.  Because of this, we only see the PCA projection of the lines plotted in Figure \ref{fig:decision_bound_lines} (a), and hence do not see such a sharp decision boundary as we do in \ref{fig:decision_boundary} (b).

\subsection{Additional views from the SharpShooter process}
\begin{figure}[!htpb]
    \centering
    \includegraphics[width=\linewidth]{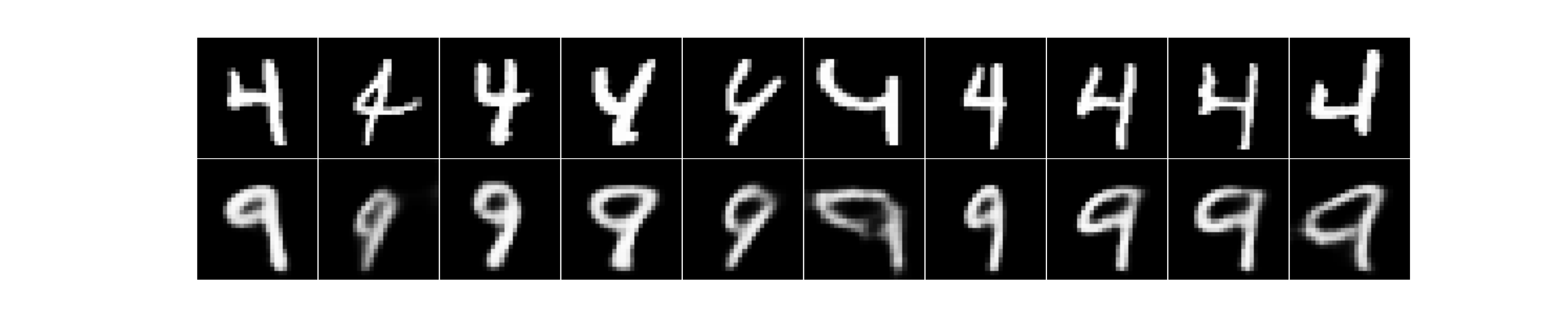}
    \caption{Demonstration of the projection qualities of the TVAE on MNIST, showing inputs (top row), and their projection (bottom row).}
    \label{fig:mnist_projections}
\end{figure}

The TVAE is responsible for projecting samples from the base class to the target class.  Philosophically, every base class sample has some innate 'goodness' to it.  The projection to the target class helps reveal those good qualities, and transforms the bad.  Figure~\ref{fig:mnist_projections} shows input samples (top) projected to the target class (bottom).  These projections on their own might be sufficient for a counterfactual explanation - however they are  typically not situated near the decision boundary.

We demonstrate counterfactual candidate generation between an input base class sample and its projection.  The candidates are generated by varying $\alpha$ in the range \begin{math}(0,1]\end{math}. A candidate, $x'$, is chosen as the counterfactual sample, $x_{cf}$, if it classifies as the target class and is within a user specified tolerance of the decision boundary.

\begin{figure}[!htpb]
    \centering
    \includegraphics[width=\linewidth]{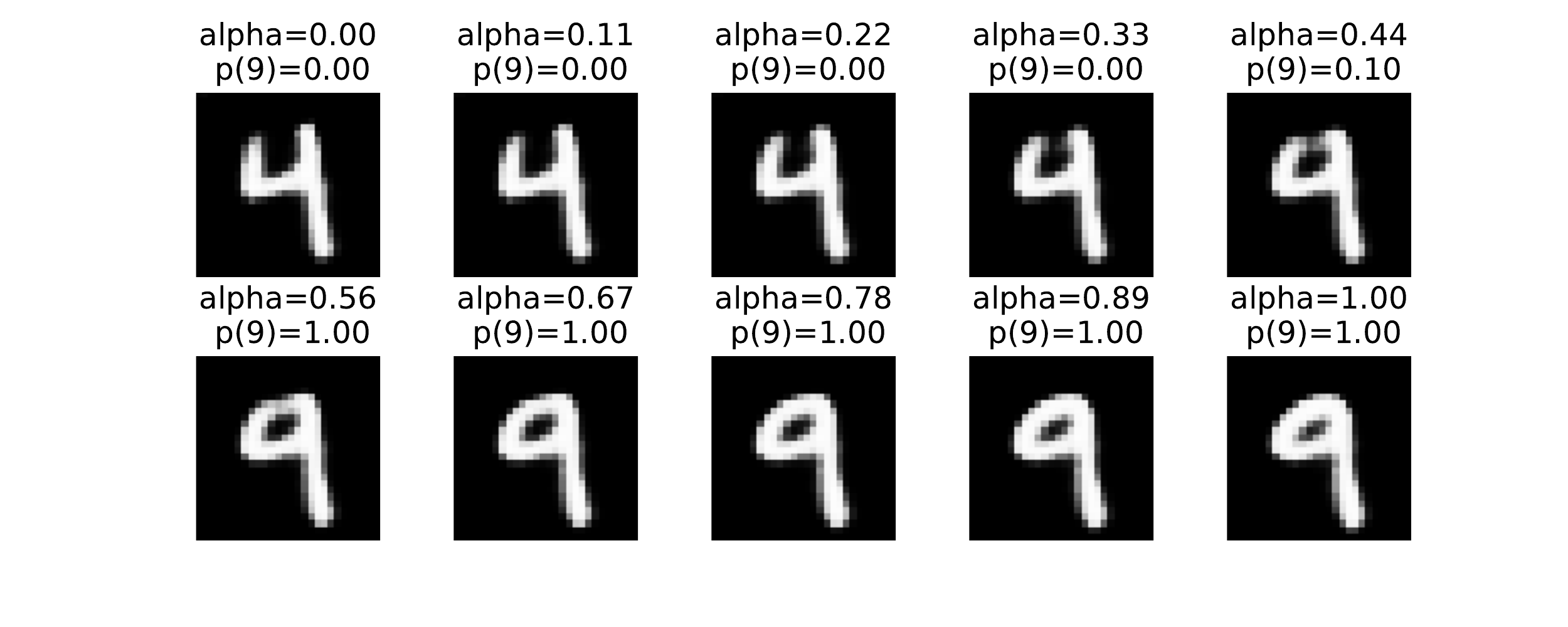}

    \caption{Counterfactual candidate generation over uniformly sampled interpolation coefficients, $\alpha$, from the base class input (left), to its target class projection (right) along with the classifier probability of being a nine. MNIST counterfactual candidates along the line see steep gradients in classifier predictions (between $\alpha=0.44$ and $\alpha=0.56$ - unseen in the tabular experiments.}
    \label{fig:mnist_interpolation}
\end{figure}

\subsection{VAE hyperparameter tuning}
For each dataset, two separate variational autoencoders were trained: one trained only on the target class (TVAE) and one that was trained on both classes (UVAE). The TVAE and the UVAE had the same architecture and only differed in hyperparameters which will be discussed for each data set below.  

For the tabular datasets, UCI credit and Lending Club, we applied a weight to influence the fidelity of reconstruction of the  categorical and continuous variables.  This weight was applied in the loss function, a combination of mean squared error and categorical or binary cross entropy for UCI credit and Lending Club respectively.  The categorical weight was subject to a hyperparameter search over the range of zero to one.

\begin{figure}[!htpb]
     \centering
     \includegraphics[width=.98\linewidth]{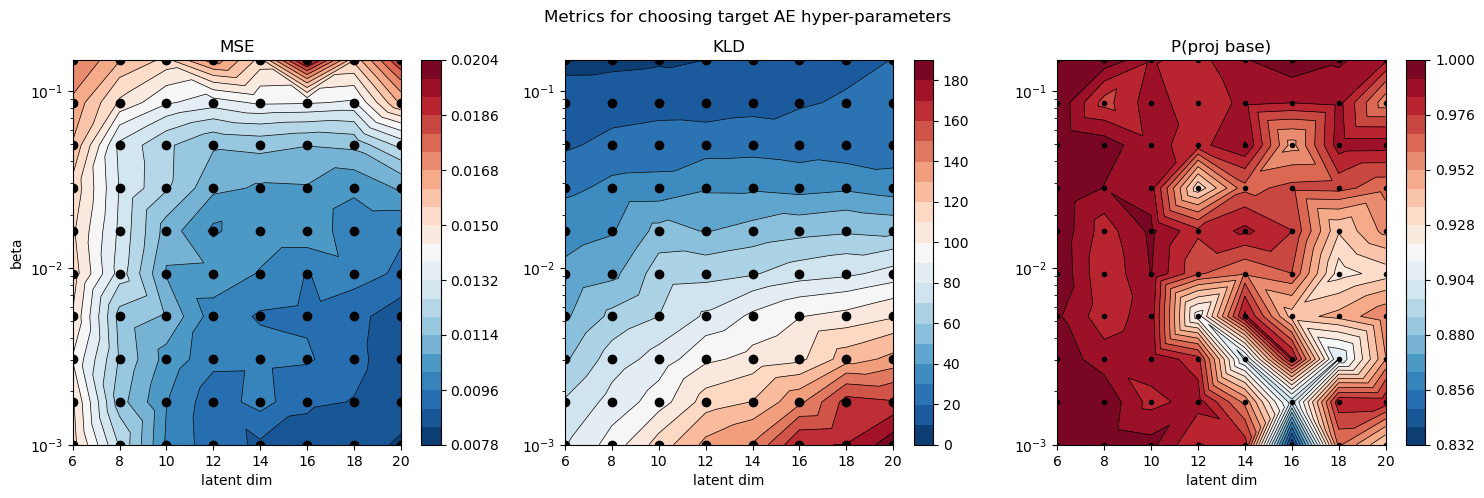}
     \caption{Contours of reconstruction error (left), KL divergence (middle), and average probability of projected base class samples (right) over a range of latent dimension and $KL$ weight for the MNIST TVAE.}
     \label{fig:mnist_grid_search_tvae}
 \end{figure}

\begin{figure}[!htpb]
     \centering
     \includegraphics[width=.98\textwidth]{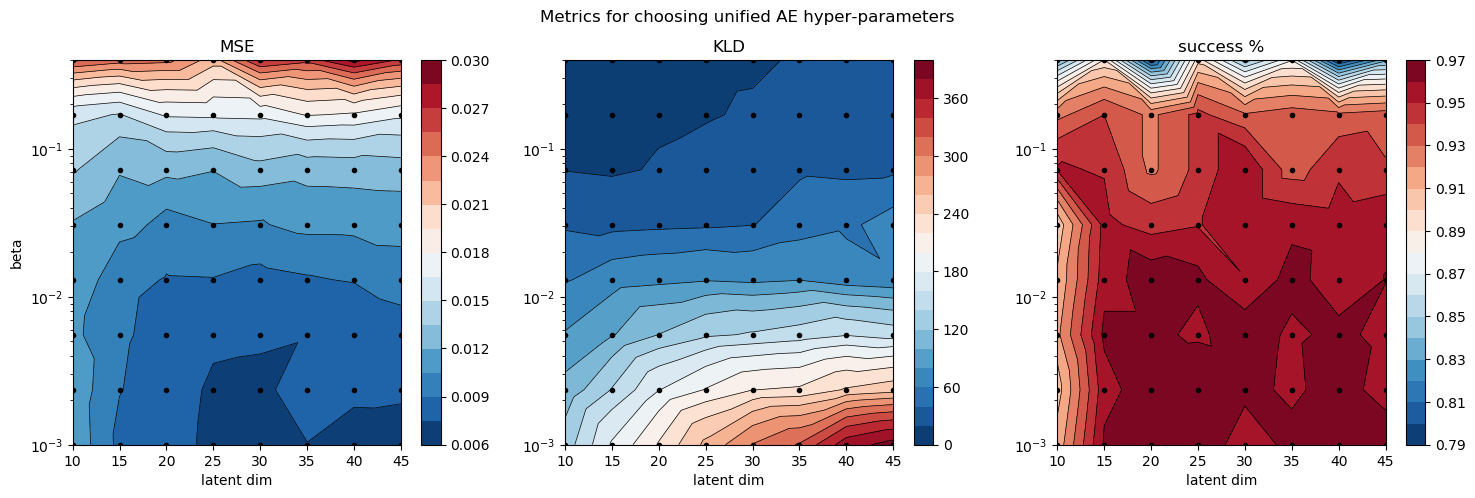}
     \caption{Contours of reconstruction error (left), KL divergence (middle), and average probability of projected base class samples (right) over a range of latent dimension and $KL$ weight for the MNIST UVAE.}
     \label{fig:mnist_grid_search_uvae}
 \end{figure}

\paragraph{MNIST}
 For MNIST, the TVAE and UVAE differed on $KL$ weight and latent dimension.  The objective is to have good reconstruction, while maintaining generative capability of the autoencoders (i.e. low mean squared error and low KL divergence).  A tertiary consideration for the TVAE was the average probability of the projected base class samples - but was not used in the decision process.  Contours of these objectives can be seen in Figure~\ref{fig:mnist_grid_search_tvae} for the TVAE, and in Figure~\ref{fig:mnist_grid_search_uvae} for the UVAE.

 The decision is brought into sharper focus by looking at the Pareto plot of the two primary objectives.  The typical \emph{elbow} shape is seen - the optimal points are lower and left most, where you cannot improve one objective without hurting the other.

\begin{figure*}[ht!]
    \centering
    \begin{subfigure}{0.49\textwidth}
        \centering
        \includegraphics[width=.95\textwidth]{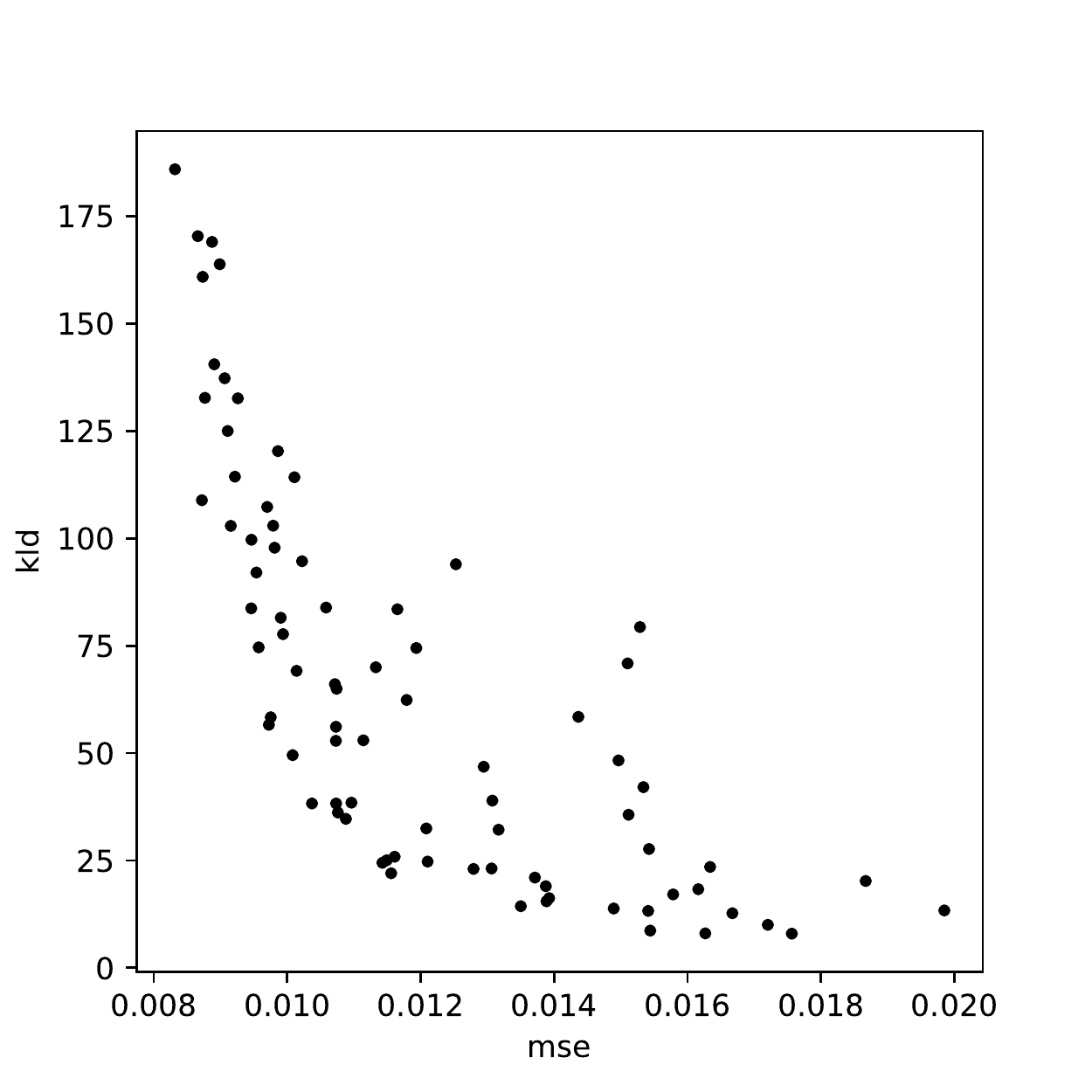}
        \caption{MNIST TVAE}
    \end{subfigure}
    \begin{subfigure}{0.49\textwidth}
        \centering
        \includegraphics[width=.95\textwidth]{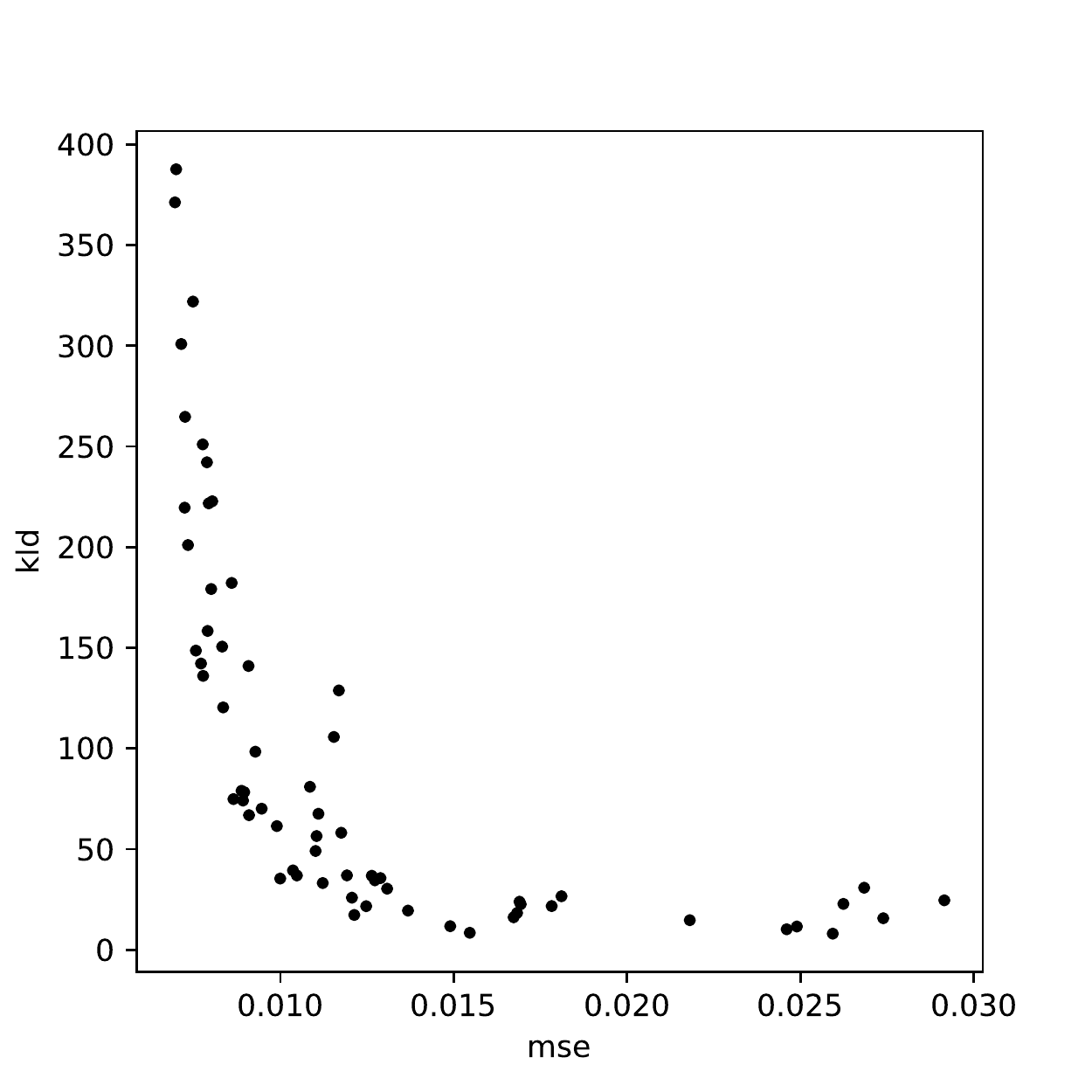}
        \caption{MNIST UVAE}
    \end{subfigure}
    
    \caption{Pareto plot of reconstruction error (mse) versus KL divergence (kld) for the VAE hyperparameter search for the MNIST dataset.  Optimal choices occur on the Pareto front where no single objective can be improved without making the other objective worse off.} 
    \label{fig:mnist_pareto}
\end{figure*}

The final choices of parameters for the TVAE were: $KL$ weight of 0.085961 with a latent dimension of fourteen.  The final choice of hyperparameters for the UVAE were: $KL$ weight of 0.013037 and a latent dimension of twenty.

\begin{figure}[!htpb]
    \centering
    \includegraphics[width=0.98\linewidth]{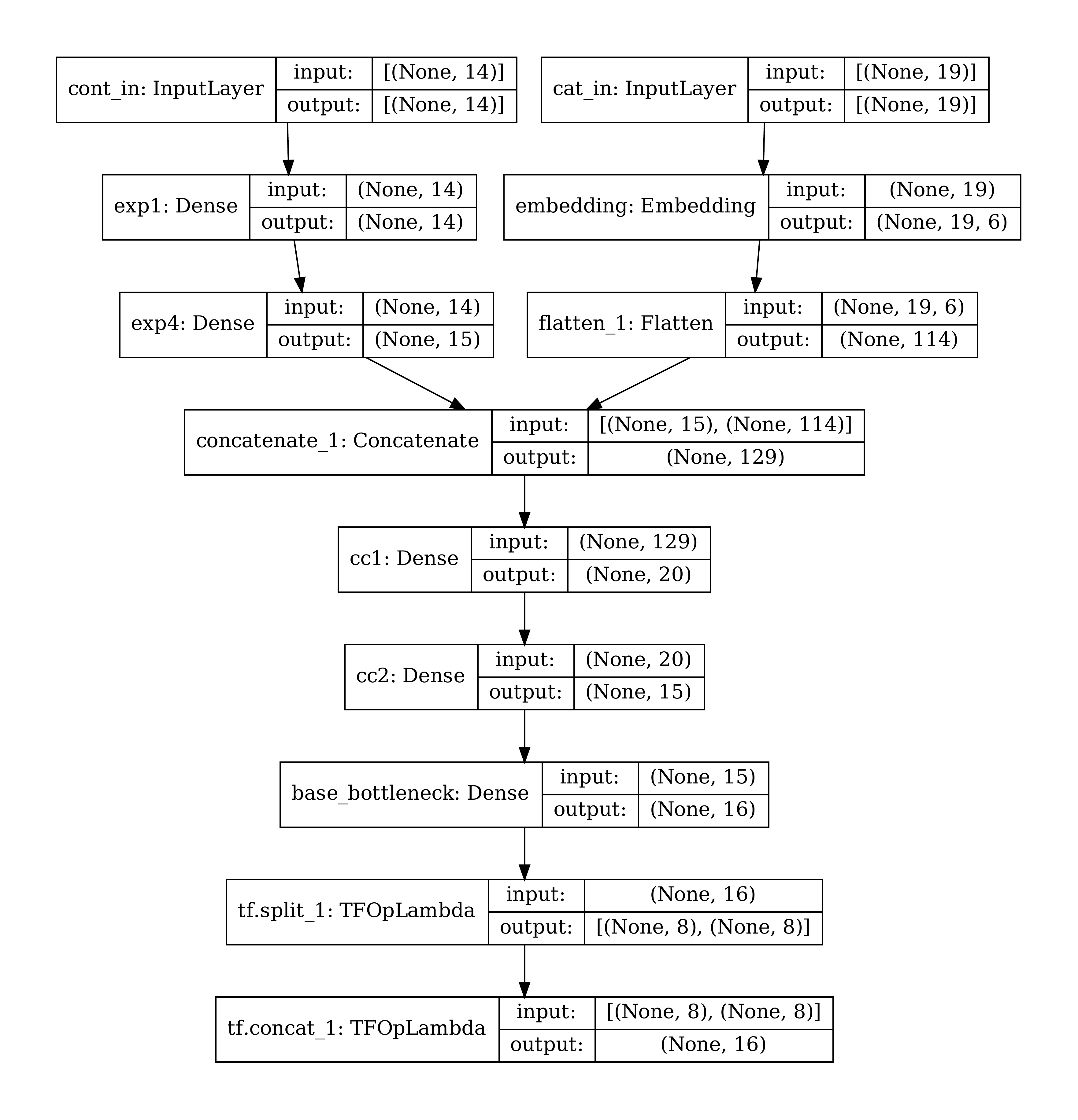}
    \caption{Architecture of the UCI credit TVAE encoder. Shows similar \emph{cat2vec} structure to merge continuous and categorical features.}
    \label{fig:UCI_tvae_encoder}
\end{figure}

\paragraph{UCI credit}
For the UCI credit dataset we used an architecture that joined together categorical and continuous variables after passing both sets through separate branches of dense or embedding layers (for the categorical and continuous variables respectively).  The two branches were then combined again with a few dense layers before the bottleneck layer, see Figure~\ref{fig:UCI_tvae_encoder}.  This process was then reversed in the decoder and is motivated by recent work on embeddings~\cite{cat2vec}.

\begin{figure}[!htpb]
     \centering
     \includegraphics[width=.98\linewidth]{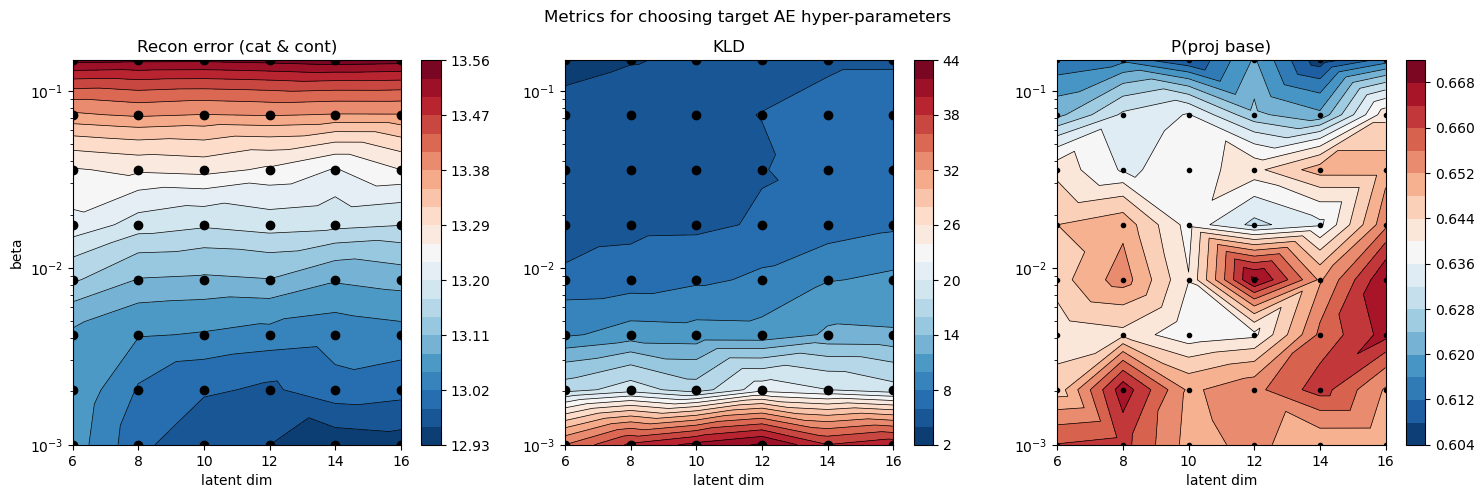}
     \caption{Contours of reconstruction error (left), KL divergence (middle), and average probability of projected base class samples (right) over a range of latent dimension and $KL$ weight for the MNIST TVAE.}
     \label{fig:uci_grid_search_tvae}
 \end{figure}
 
 \begin{figure}[!htpb]
     \centering
     \includegraphics[width=.98\textwidth]{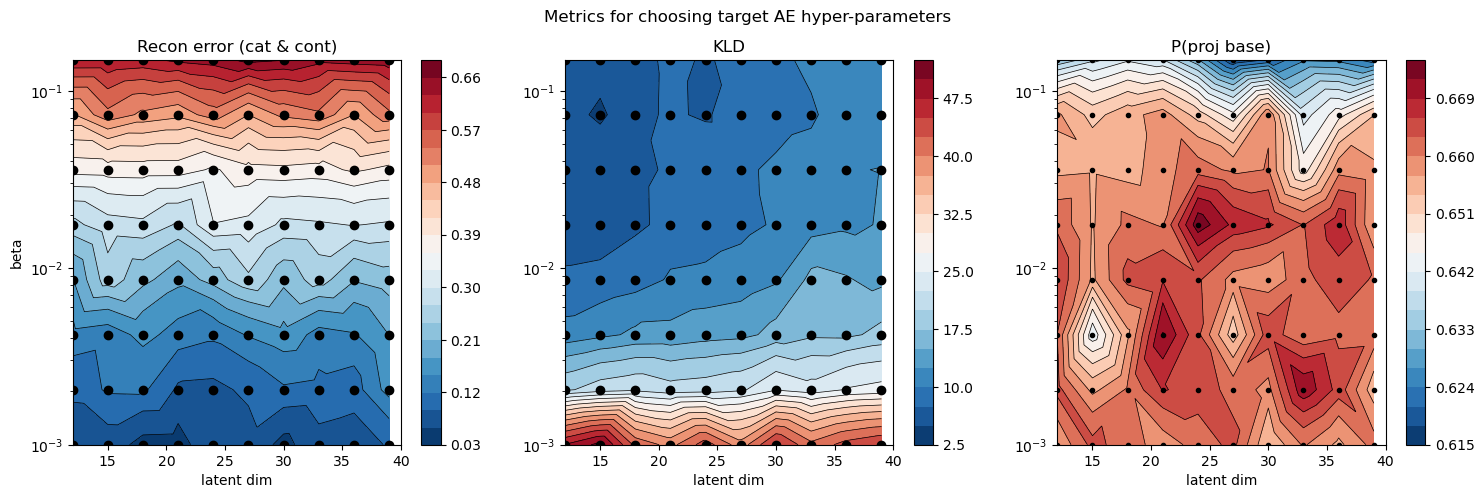}
     \caption{Contours of reconstruction error (left), KL divergence (middle), and average probability of projected base class samples (right) over a range of latent dimension and $KL$ weight for the MNIST UVAE.}
     \label{fig:uci_grid_search_uvae}
 \end{figure}

\paragraph{Lending Club}

\begin{figure}[!htpb]
     \centering
     \includegraphics[width=.98\linewidth]{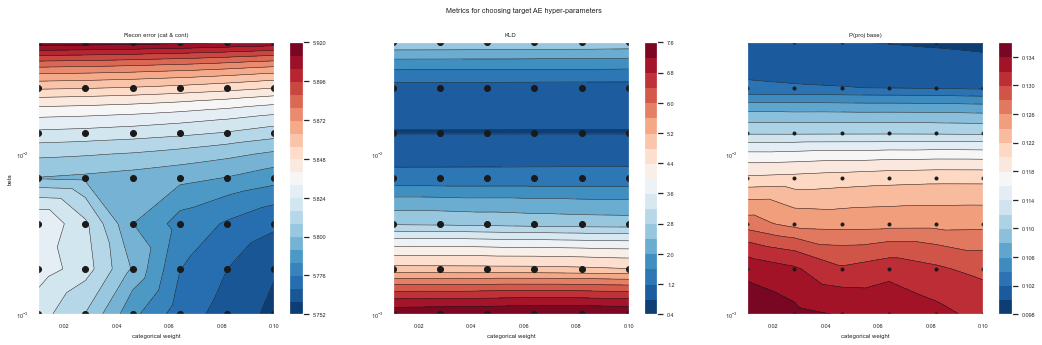}
     \caption{Contours of reconstruction loss (left), $KL$ divergence (middle), and average probability of the projected base samples in the test set (right) for the Lending Club TVAE.}
     \label{fig:lc_target_grid_search}
 \end{figure}

For Lending Club, with only 6 input features, the latent dimension was fixed to be two.  The difference in hyperparameters occurred for $KL$ weight and categorical weight used to combine the means squared error of the continuous variables with the binary cross entropy of the categorical.   These hyperparameters were put to a grid search and the final set were chosen in the same manner as the MNIST  hyperparameters. The contours of reconstruction error and KL divergence are shown in Figure~\ref{fig:lc_target_grid_search}. 

Each objective has implications on the resulting counterfactuals. For instance, a high categorical weight tended to pull apart the latent space and would make linear interpolation less feasible, such as in the UCI dataset. Candidates being accepted in areas of low probability space in $p(z|x)$ could imply resulting counterfactuals with low mass in $p(x)$ and could be considered less feasible. Similarly, a higher $KL$ weights help encourage convexity through a "regularization" of the codes towards their standard normal prior.  Convexity could be advantageous to SharpShooter by again preventing traversing latent space with low probability mass, but also empirically it seemed to provide a more defined separation of the classes. Finally, reconstruction quality was important due to the desire for base observations to be close to their projections and their interpolated candidates.

 \begin{figure}[!htpb]
     \centering
     \includegraphics[width=.98\linewidth]{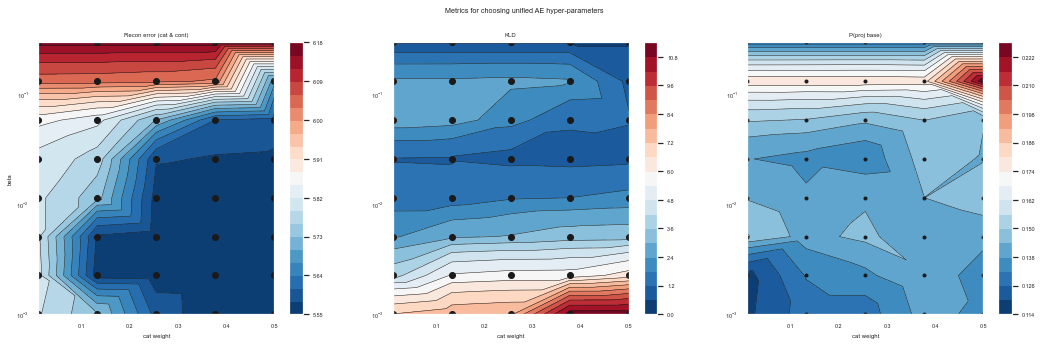}
     \caption{Contours of reconstruction loss (left), $KL$ divergence (middle), and average probability of the projected base samples in the test set (right) for the Lending Club UVAE.}
     \label{fig:lc_uvae_grid_search}
 \end{figure}

  The algorithm would look like this:
\begin{algorithm}[h]
\Parameter{$p$ probability of target counterfactual class (0.5 for decision boundary), $tol$ tolerance}
\KwIn{Classifier $f$, VAE trained on $X_t$, VAE trained on $(X_b, X_t)$, $\alpha \in (0,1)$} 
\KwOut{$X_{cf}$ a counterfactual}

$\hat{X}_t  = \phi_t(X_b)$ \\  
$Z_b = \varphi_{b,t}(X_b)$ \\  
$\hat{Z}_t = \varphi_{b,t}(\hat{X}_t)$ \\
Initialize $X' = X_b$

\While{$\abs{f(X') - T > tol}$ \textbf{and} $f(X') < p$}{
    $Z' = \alpha \hat{Z}_t + (1-\alpha) Z_b$\\
    $X' = \psi_{b,t}(Z')$ \\
    $\mathcal{L}(\alpha) = ||p - f(X')|| $ \\
    $\alpha = \alpha + \eta \nabla_\alpha \mathcal{L} $
}
\Return $X_{cf} =X'$
\caption{SharpShooter Gradient Descent}
\label{algo:GD}
\end{algorithm}

\subsection{Classifier architecture}
SharpShooter is agnostic to the architecture of the classifier used.  We include the architecture of the classifiers for repeatability.  The classifiers used for this study were all built in the Keras API of Tensorflow.  Model architecture diagrams are show for each data set below.

\begin{figure}[!htpb]
    \centering
    \includegraphics[width=0.5\linewidth]{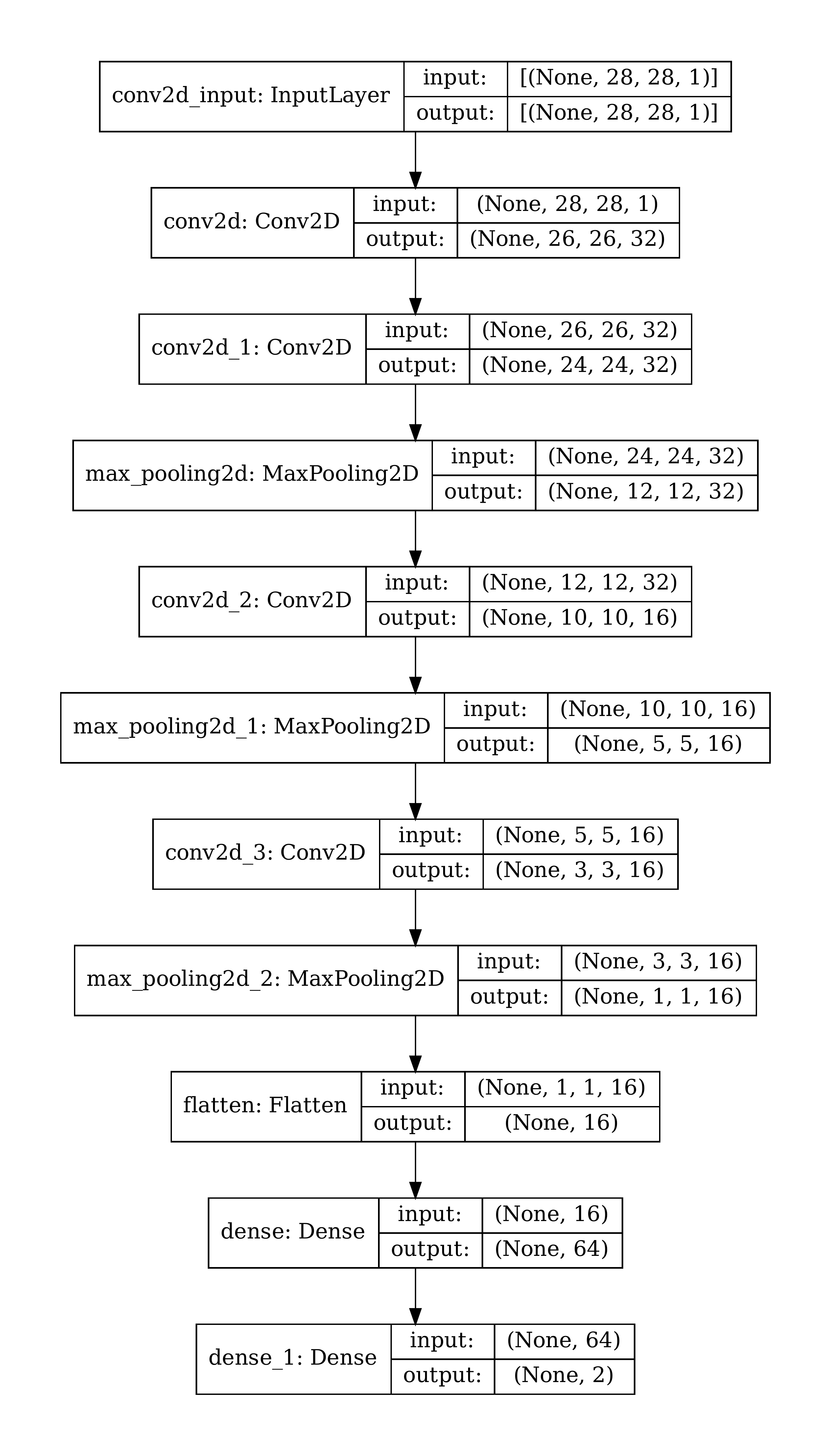}
    \caption{Architecture of the MNIST classifier.}
    \label{fig:mnist_classifier}
\end{figure}

\begin{figure}[!htpb]
    \centering
    \includegraphics[width=0.75\linewidth]{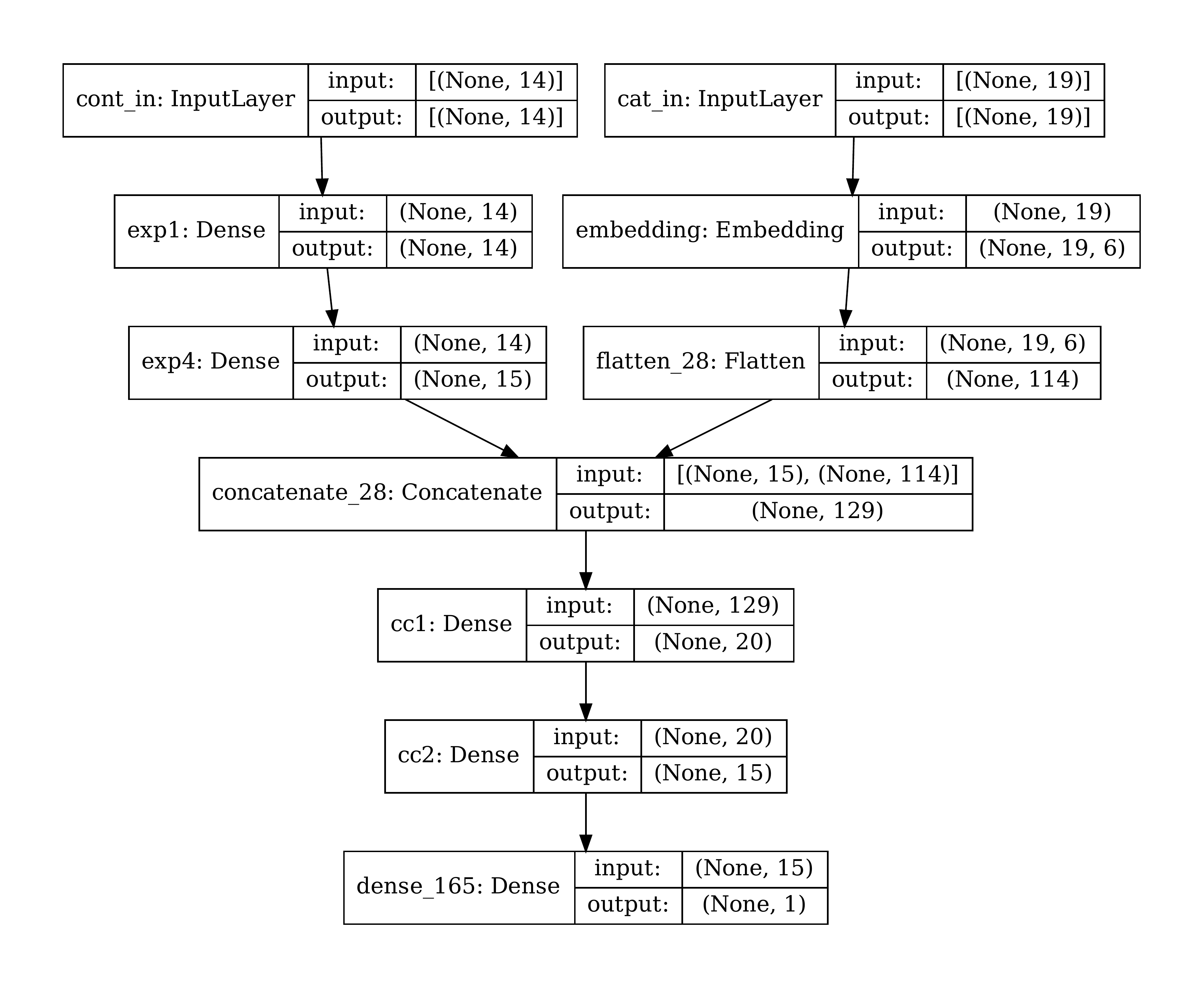}
    \caption{Architecture of the UCI classifier - uses a \emph{cat2vec} approach to learn an embedding from the categorical features before concatenating with the dense layers used with the continuous variables..}
    \label{fig:uci_classifier}
\end{figure}

For the Lending Club dataset, we down selected to six features; the classifier achieved an AUC of $0.96$.  

\begin{figure}[!htpb]
    \centering
    \includegraphics[width=0.5\linewidth]{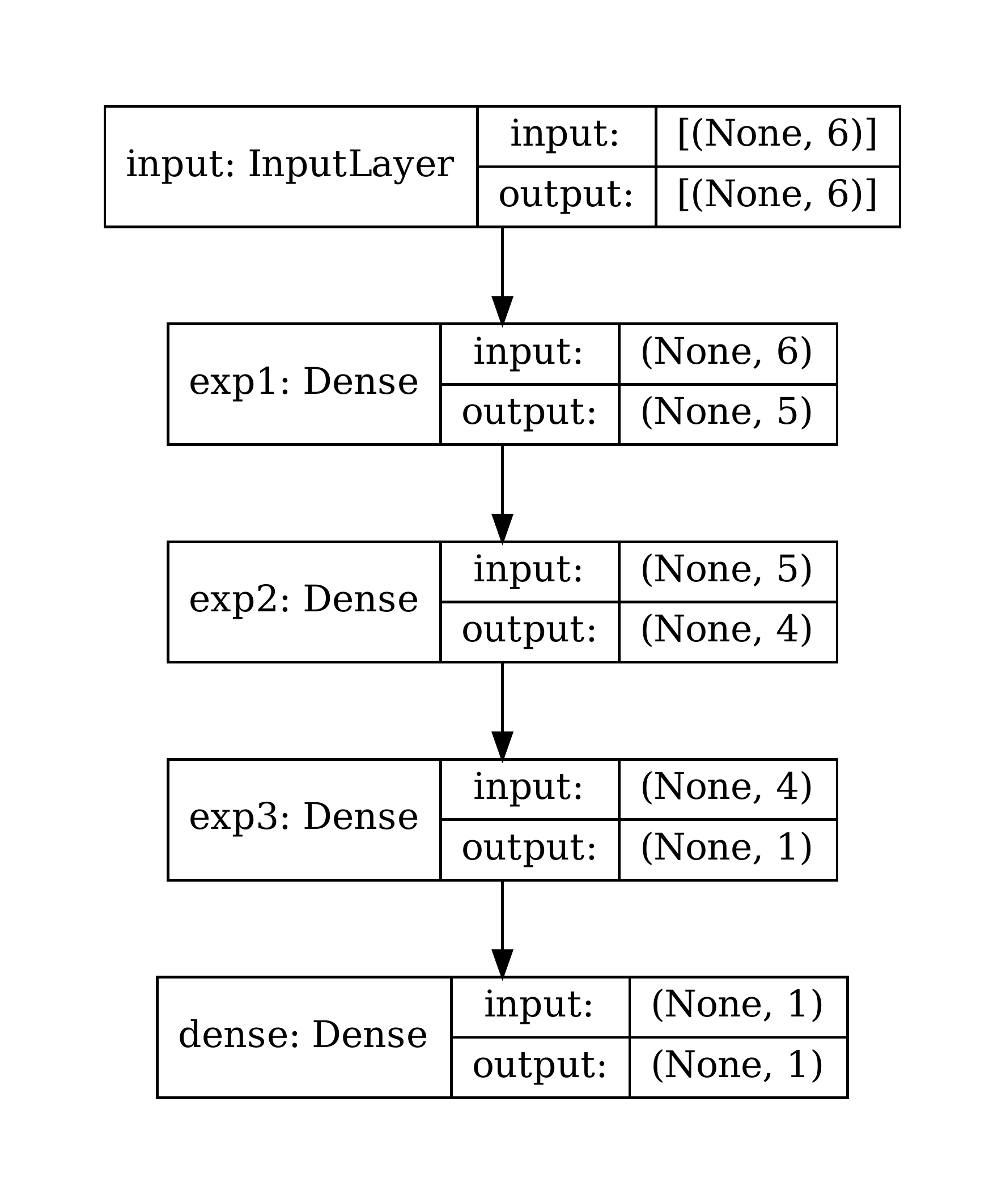}
    \caption{Architecture of the Lending Club classifier.}
    \label{fig:lc_classifier}
\end{figure}

\subsection{UCI VAE architecture}
 The combination of continuous and categorical features for the UCI credit dataset required an interesting architecture, that we show below.
 
\begin{figure}[!htpb]
    \centering
    \includegraphics[width=0.98\linewidth]{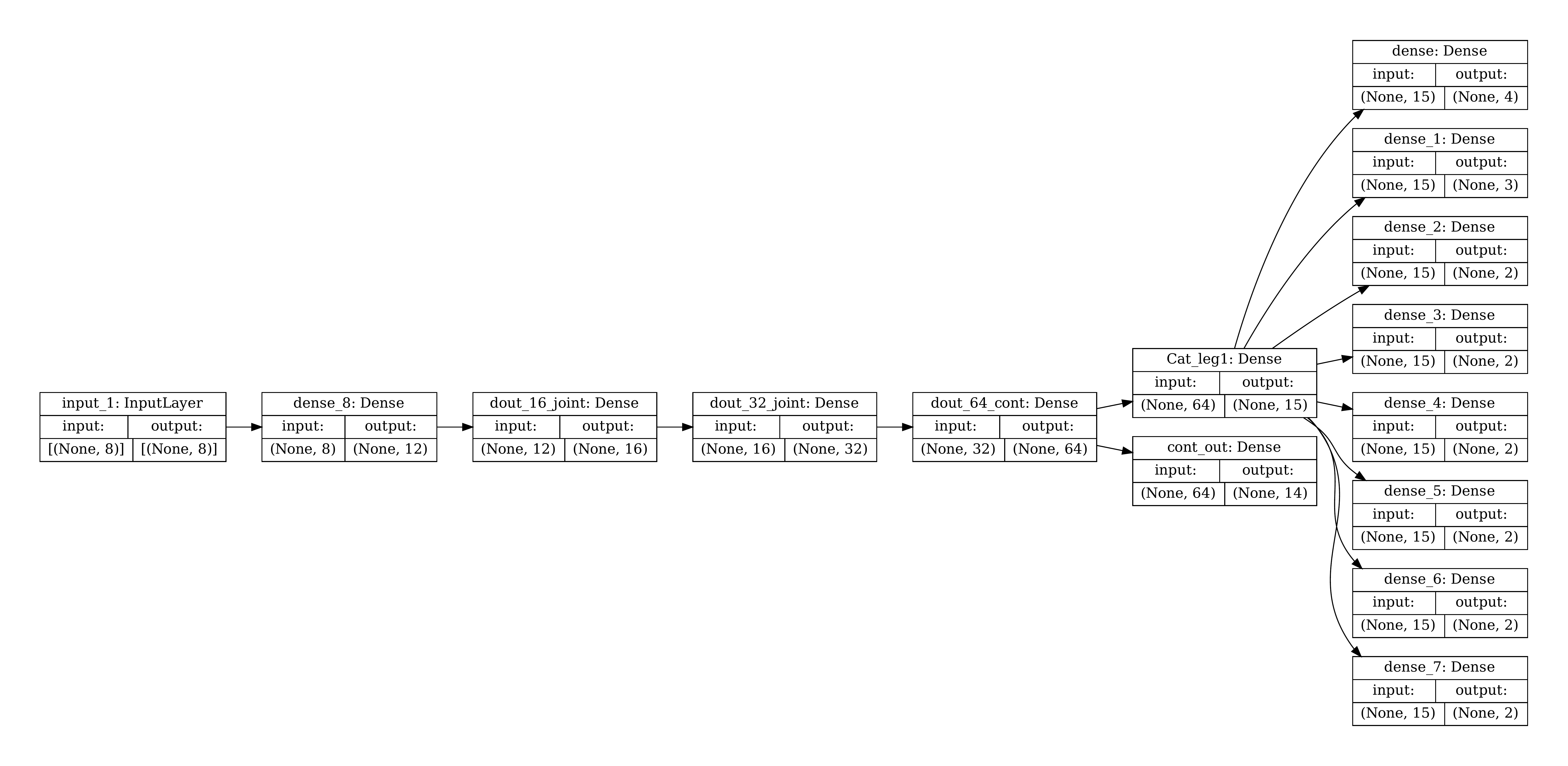}
    \caption{Architecture of the UCI credit TVAE encoder. Oriented from left (input) to right (output).}
    \label{fig:UCI_tvae_decoder}
\end{figure}

For the decoder, the categorical features are split off into distinct outputs, shown on the right in Figure~\ref{fig:UCI_tvae_decoder}, from the single continuous output, labeled 'cont\_out' in the figure.

\end{document}